%% file: main_arxiv.tex
\documentclass{article}

\usepackage{styling/arxiv}

\usepackage[utf8]{inputenc} 
\usepackage[T1]{fontenc}    
\usepackage{hyperref}       
\usepackage{url}            
\usepackage{booktabs}       
\usepackage{amsfonts}       
\usepackage{nicefrac}       
\usepackage{microtype}      
\usepackage{lipsum}		
\usepackage{graphicx}
\usepackage{doi}
\usepackage{bm}
\usepackage{amsmath}
\usepackage{amssymb}
\usepackage{amsthm}
\usepackage{algorithm}
\usepackage{cite}
\usepackage{arydshln}
\usepackage{algpseudocode}
\usepackage{setspace}
\newtheorem{Theorem}{Theorem}[section]

\usepackage{caption}
\usepackage{subcaption}

\newif\ifEntropy
\Entropyfalse

\usepackage[]{todonotes}   

\input{content/00-preamble}

\title{Automating Model Comparison in Factor Graphs}

\date{\today}

\author{
  Bart van Erp$^{1,}$\thanks{Correspondence: \href{mailto:b.v.erp@tue.nl}{b.v.erp@tue.nl}} \qquad Wouter W. L. Nuijten$^1$ \qquad Thijs van de Laar$^1$ \qquad Bert de Vries$^{1, 2}$ \\
  $^{1}$Department of Electrical Engineering, Eindhoven University of Technology, Eindhoven, The Netherlands \\
  $^{2}$GN Hearing, Eindhoven, The Netherlands
}

\renewcommand{\shorttitle}{Automating Model Comparison in Factor Graphs}

\begin{document}
\maketitle

\setcounter{footnote}{0}

\input{content/01-abstract}
\input{content/02-introduction}
\input{content/03-related}
\input{content/04-background}
\input{content/05-mixture}
\input{content/06-comparison}

\input{content/07-experiments}
\input{content/08-discussion}
\input{content/09-conclusion}

\section*{Acknowledgments}
This work was partly financed by GN Hearing A/S. 
The authors would like to thank the \href{https://biaslab.github.io/}{BIASlab} team members for various insightful discussions related to this work.

\bibliographystyle{IEEEtran}
\bibliography{content/references_western}

\appendix
\newpage
\input{content/A1-proofs}
\input{content/A2-derivations}
\end{document}

%% file: content/00-preamble.tex
\usepackage{pgfplots}
\pgfplotsset{compat=1.17}
\usetikzlibrary{intersections}
\usepackage{bm}
\usetikzlibrary{positioning}
\usepgfplotslibrary{fillbetween}
\usepgfplotslibrary{groupplots}
\usetikzlibrary{shapes.geometric,backgrounds}
\usepackage[outline]{contour}

\usepackage{tikz}
\usepackage{xifthen}

\pgfdeclarelayer{bg}    
\pgfsetlayers{bg,main}  
\definecolor{beige}{RGB}{245, 245, 220}

\definecolor{darkgrey}{RGB}{75, 75, 75}
\definecolor{lightgrey}{RGB}{250, 250, 250}

\usetikzlibrary{calc, arrows, fit, positioning, patterns, decorations.pathreplacing, shapes}
\tikzstyle{dash} = [dashed, -latex,>=latex]
\tikzstyle{line} = [draw, -latex,>=latex]
\tikzstyle{smallbox} = [draw, minimum size=5.0mm]
\tikzstyle{box} = [draw, minimum size=7.0mm]
\tikzstyle{bigbox} = [draw, minimum size=10.0mm]
\tikzstyle{switch} = [trapezium, trapezium angle=120, draw, rotate=90,  inner ysep=5pt, outer sep=5pt,
minimum height=7mm, minimum width=7mm]
\tikzstyle{roundbox} = [draw, circle, inner sep=0pt, minimum size=3mm]
\tikzstyle{clamped} = [draw, fill=black, minimum size=0.15cm]
\tikzstyle{msgcircle} = [shape=circle, draw, inner sep=0pt, minimum size=4mm, fill=white, font=\scriptsize]
\tikzstyle{darkmsgcircle} = [shape=circle, draw, inner sep=0pt, minimum size=4mm, fill=darkgrey, text=white, font=\scriptsize]
\tikzstyle{msgdoublecircle} = [shape=circle, double, double distance=1.5pt, draw, inner sep=0pt, minimum size=5mm, fill=white]
\tikzstyle{darkmsgdoublecircle} = [shape=circle, double, double distance=1.5pt, draw, inner sep=0pt, minimum size=5mm, fill=darkgrey, text=white, font=\bfseries]

\newcommand{\msg}[6]{
      \ifthenelse{\isin{#1}{left} \AND \isin{#2}{down}}{
            \coordinate (anchor) at ($({#3})!{#5}!({#4})$);
            \node[xshift=-6.0mm] at (anchor) {#6};
            \node[xshift=-1.0mm] at (anchor) {$\downarrow$};
      }{}
      \ifthenelse{\isin{#1}{right} \AND \isin{#2}{down}}{
            \coordinate (anchor) at ($({#3})!{#5}!({#4})$);
            \node[xshift=6.0mm] at (anchor) {#6};
            \node[xshift=1.0mm] at (anchor) {$\downarrow$};
      }{}

      \ifthenelse{\isin{#1}{down} \AND \isin{#2}{right}}{
            \coordinate (anchor) at ($({#3})!{#5}!({#4})$);
            \node[ yshift=-4.0mm] at (anchor) {#6};
            \node[yshift=-1.0mm] at (anchor) {$\rightarrow$};
      }{}
      \ifthenelse{\isin{#1}{up} \AND \isin{#2}{right}}{
            \coordinate (anchor) at ($({#3})!{#5}!({#4})$);
            \node[ yshift=4.0mm] at (anchor) {#6};
            \node[yshift=1.0mm] at (anchor) {$\rightarrow$};
      }{}

      \ifthenelse{\isin{#1}{down} \AND \isin{#2}{left}}{
            \coordinate (anchor) at ($({#3})!{#5}!({#4})$);
            \node[ yshift=-4.0mm] at (anchor) {#6};
            \node[yshift=-1.0mm] at (anchor) {$\leftarrow$};
      }{}
      \ifthenelse{\isin{#1}{up} \AND \isin{#2}{left}}{
            \coordinate (anchor) at ($({#3})!{#5}!({#4})$);
            \node[ yshift=4.0mm] at (anchor) {#6};
            \node[yshift=1.0mm] at (anchor) {$\leftarrow$};
      }{}

      \ifthenelse{\isin{#1}{left} \AND \isin{#2}{up}}{
            \coordinate (anchor) at ($({#3})!{#5}!({#4})$);
            \node[ xshift=-6.0mm] at (anchor) {#6};
            \node[xshift=-1.0mm] at (anchor) {$\uparrow$};
      }{}
      \ifthenelse{\isin{#1}{right} \AND \isin{#2}{up}}{
            \coordinate (anchor) at ($({#3})!{#5}!({#4})$);
            \node[ xshift=6.0mm] at (anchor) {#6};
            \node[xshift=1.0mm] at (anchor) {$\uparrow$};
      }{}
}

\newcommand{\msgcircle}[6]{
      \ifthenelse{\isin{#1}{left} \AND \isin{#2}{down}}{
            \coordinate (anchor) at ($({#3})!{#5}!({#4})$);
            \node[msgcircle,xshift=-5.0mm] at (anchor) {#6};
            \node[xshift=-1.5mm] at (anchor) {$\downarrow$};
      }{}
      \ifthenelse{\isin{#1}{right} \AND \isin{#2}{down}}{
            \coordinate (anchor) at ($({#3})!{#5}!({#4})$);
            \node[msgcircle,xshift=5.0mm] at (anchor) {#6};
            \node[xshift=1.5mm] at (anchor) {$\downarrow$};
      }{}

      \ifthenelse{\isin{#1}{down} \AND \isin{#2}{right}}{
            \coordinate (anchor) at ($({#3})!{#5}!({#4})$);
            \node[msgcircle, yshift=-5.0mm] at (anchor) {#6};
            \node[yshift=-2.0mm] at (anchor) {$\rightarrow$};
      }{}
      \ifthenelse{\isin{#1}{up} \AND \isin{#2}{right}}{
            \coordinate (anchor) at ($({#3})!{#5}!({#4})$);
            \node[msgcircle, yshift=5.0mm] at (anchor) {#6};
            \node[yshift=2.0mm] at (anchor) {$\rightarrow$};
      }{}

      \ifthenelse{\isin{#1}{down} \AND \isin{#2}{left}}{
            \coordinate (anchor) at ($({#3})!{#5}!({#4})$);
            \node[msgcircle, yshift=-5.0mm] at (anchor) {#6};
            \node[yshift=-2.0mm] at (anchor) {$\leftarrow$};
      }{}
      \ifthenelse{\isin{#1}{up} \AND \isin{#2}{left}}{
            \coordinate (anchor) at ($({#3})!{#5}!({#4})$);
            \node[msgcircle, yshift=5.0mm] at (anchor) {#6};
            \node[yshift=2.0mm] at (anchor) {$\leftarrow$};
      }{}

      \ifthenelse{\isin{#1}{left} \AND \isin{#2}{up}}{
            \coordinate (anchor) at ($({#3})!{#5}!({#4})$);
            \node[msgcircle, xshift=-5.0mm] at (anchor) {#6};
            \node[xshift=-1.5mm] at (anchor) {$\uparrow$};
      }{}
      \ifthenelse{\isin{#1}{right} \AND \isin{#2}{up}}{
            \coordinate (anchor) at ($({#3})!{#5}!({#4})$);
            \node[msgcircle, xshift=5.0mm] at (anchor) {#6};
            \node[xshift=1.5mm] at (anchor) {$\uparrow$};
      }{}
}

\newcommand{\darkmsg}[6]{
      \ifthenelse{\isin{#1}{left} \AND \isin{#2}{down}}{
            \coordinate (anchor) at ($({#3})!{#5}!({#4})$);
            \node[darkmsgcircle, xshift=-5mm] at (anchor) {#6};
            \node[xshift=-1.5mm] at (anchor) {$\downarrow$};
      }{}
      \ifthenelse{\isin{#1}{right} \AND \isin{#2}{down}}{
            \coordinate (anchor) at ($({#3})!{#5}!({#4})$);
            \node[darkmsgcircle, xshift=5mm] at (anchor) {#6};
            \node[xshift=1.5mm] at (anchor) {$\downarrow$};
      }{}

      \ifthenelse{\isin{#1}{down} \AND \isin{#2}{right}}{
            \coordinate (anchor) at ($({#3})!{#5}!({#4})$);
            \node[darkmsgcircle, yshift=-5.0mm] at (anchor) {#6};
            \node[yshift=-2.0mm] at (anchor) {$\rightarrow$};
      }{}
      \ifthenelse{\isin{#1}{up} \AND \isin{#2}{right}}{
            \coordinate (anchor) at ($({#3})!{#5}!({#4})$);
            \node[darkmsgcircle, yshift=5.0mm] at (anchor) {#6};
            \node[yshift=2.0mm] at (anchor) {$\rightarrow$};
      }{}

      \ifthenelse{\isin{#1}{down} \AND \isin{#2}{left}}{
            \coordinate (anchor) at ($({#3})!{#5}!({#4})$);
            \node[darkmsgcircle, yshift=-5.0mm] at (anchor) {#6};
            \node[yshift=-2.0mm] at (anchor) {$\leftarrow$};
      }{}
      \ifthenelse{\isin{#1}{up} \AND \isin{#2}{left}}{
            \coordinate (anchor) at ($({#3})!{#5}!({#4})$);
            \node[darkmsgcircle, yshift=5.0mm] at (anchor) {#6};
            \node[yshift=2.0mm] at (anchor) {$\leftarrow$};
      }{}

      \ifthenelse{\isin{#1}{left} \AND \isin{#2}{up}}{
            \coordinate (anchor) at ($({#3})!{#5}!({#4})$);
            \node[darkmsgcircle, xshift=-5.0mm] at (anchor) {#6};
            \node[xshift=-1.5mm] at (anchor) {$\uparrow$};
      }{}
      \ifthenelse{\isin{#1}{right} \AND \isin{#2}{up}}{
            \coordinate (anchor) at ($({#3})!{#5}!({#4})$);
            \node[darkmsgcircle, xshift=5.0mm] at (anchor) {#6};
            \node[xshift=1.5mm] at (anchor) {$\uparrow$};
      }{}
}

\newcommand{\bwmsg}[6]{
      \ifthenelse{\isin{#1}{left} \AND \isin{#2}{down}}{
            \coordinate (anchor) at ($({#3})!{#5}!({#4})$);
            \node[msgdoublecircle, xshift=-5.5mm] at (anchor) {#6};
            \node[xshift=-1.5mm] at (anchor) {$\downarrow$};
      }{}
      \ifthenelse{\isin{#1}{right} \AND \isin{#2}{down}}{
            \coordinate (anchor) at ($({#3})!{#5}!({#4})$);
            \node[msgdoublecircle, xshift=5.5mm] at (anchor) {#6};
            \node[xshift=1.5mm] at (anchor) {$\downarrow$};
      }{}

      \ifthenelse{\isin{#1}{down} \AND \isin{#2}{right}}{
            \coordinate (anchor) at ($({#3})!{#5}!({#4})$);
            \node[msgdoublecircle, yshift=-6.0mm] at (anchor) {#6};
            \node[yshift=-2.0mm] at (anchor) {$\rightarrow$};
      }{}
      \ifthenelse{\isin{#1}{up} \AND \isin{#2}{right}}{
            \coordinate (anchor) at ($({#3})!{#5}!({#4})$);
            \node[msgdoublecircle, yshift=6.0mm] at (anchor) {#6};
            \node[yshift=2.0mm] at (anchor) {$\rightarrow$};
      }{}

      \ifthenelse{\isin{#1}{down} \AND \isin{#2}{left}}{
            \coordinate (anchor) at ($({#3})!{#5}!({#4})$);
            \node[msgdoublecircle, yshift=-6.0mm] at (anchor) {#6};
            \node[yshift=-2.0mm] at (anchor) {$\leftarrow$};
      }{}
      \ifthenelse{\isin{#1}{up} \AND \isin{#2}{left}}{
            \coordinate (anchor) at ($({#3})!{#5}!({#4})$);
            \node[msgdoublecircle, yshift=6.0mm] at (anchor) {#6};
            \node[yshift=2.0mm] at (anchor) {$\leftarrow$};
      }{}

      \ifthenelse{\isin{#1}{left} \AND \isin{#2}{up}}{
            \coordinate (anchor) at ($({#3})!{#5}!({#4})$);
            \node[msgdoublecircle, xshift=-5.5mm] at (anchor) {#6};
            \node[xshift=-1.5mm] at (anchor) {$\uparrow$};
      }{}
      \ifthenelse{\isin{#1}{right} \AND \isin{#2}{up}}{
            \coordinate (anchor) at ($({#3})!{#5}!({#4})$);
            \node[msgdoublecircle, xshift=5.5mm] at (anchor) {#6};
            \node[xshift=1.5mm] at (anchor) {$\uparrow$};
      }{}
}

\newcommand{\bwdarkmsg}[6]{
      \ifthenelse{\isin{#1}{left} \AND \isin{#2}{down}}{
            \coordinate (anchor) at ($({#3})!{#5}!({#4})$);
            \node[darkmsgdoublecircle, xshift=-5.5mm] at (anchor) {#6};
            \node[xshift=-1.5mm] at (anchor) {$\downarrow$};
      }{}
      \ifthenelse{\isin{#1}{right} \AND \isin{#2}{down}}{
            \coordinate (anchor) at ($({#3})!{#5}!({#4})$);
            \node[darkmsgdoublecircle, xshift=5.5mm] at (anchor) {#6};
            \node[xshift=1.5mm] at (anchor) {$\downarrow$};
      }{}

      \ifthenelse{\isin{#1}{down} \AND \isin{#2}{right}}{
            \coordinate (anchor) at ($({#3})!{#5}!({#4})$);
            \node[darkmsgdoublecircle, yshift=-6.0mm] at (anchor) {#6};
            \node[yshift=-2.0mm] at (anchor) {$\rightarrow$};
      }{}
      \ifthenelse{\isin{#1}{up} \AND \isin{#2}{right}}{
            \coordinate (anchor) at ($({#3})!{#5}!({#4})$);
            \node[darkmsgdoublecircle, yshift=6.0mm] at (anchor) {#6};
            \node[yshift=2.0mm] at (anchor) {$\rightarrow$};
      }{}

      \ifthenelse{\isin{#1}{down} \AND \isin{#2}{left}}{
            \coordinate (anchor) at ($({#3})!{#5}!({#4})$);
            \node[darkmsgdoublecircle, yshift=-6.0mm] at (anchor) {#6};
            \node[yshift=-2.0mm] at (anchor) {$\leftarrow$};
      }{}
      \ifthenelse{\isin{#1}{up} \AND \isin{#2}{left}}{
            \coordinate (anchor) at ($({#3})!{#5}!({#4})$);
            \node[darkmsgdoublecircle, yshift=6.0mm] at (anchor) {#6};
            \node[yshift=2.0mm] at (anchor) {$\leftarrow$};
      }{}

      \ifthenelse{\isin{#1}{left} \AND \isin{#2}{up}}{
            \coordinate (anchor) at ($({#3})!{#5}!({#4})$);
            \node[darkmsgdoublecircle, xshift=-5.5mm] at (anchor) {#6};
            \node[xshift=-1.5mm] at (anchor) {$\uparrow$};
      }{}
      \ifthenelse{\isin{#1}{right} \AND \isin{#2}{up}}{
            \coordinate (anchor) at ($({#3})!{#5}!({#4})$);
            \node[darkmsgdoublecircle, xshift=5.5mm] at (anchor) {#6};
            \node[xshift=1.5mm] at (anchor) {$\uparrow$};
      }{}
}

\makeatletter
\DeclareRobustCommand{\cev}[1]{%
  \mathpalette\do@cev{#1}%
}
\newcommand{\do@cev}[2]{%
  \fix@cev{#1}{+}%
  \reflectbox{$\m@th#1\vec{\reflectbox{$\fix@cev{#1}{-}\m@th#1#2\fix@cev{#1}{+}$}}$}%
  \fix@cev{#1}{-}%
}
\newcommand{\fix@cev}[2]{%
  \ifx#1\displaystyle
    \mkern#23mu
  \else
    \ifx#1\textstyle
      \mkern#23mu
    \else
      \ifx#1\scriptstyle
        \mkern#22mu
      \else
        \mkern#22mu
      \fi
    \fi
  \fi
}

\DeclareMathOperator*{\argmax}{arg\,max}

\newcommand{\midi}[0]{%
    {\,|\,}%
}

\newcommand{\diff}[0]{%
    {\,\mathrm{d}}%
}

%% file: content/01-abstract.tex
\ifEntropy
    \abstract{%
    Bayesian state and parameter estimation have been automated effectively in a variety of probabilistic programming languages.
    The process of model comparison on the other hand, which still requires error-prone and time-consuming manual derivations, is often overlooked despite its importance.
    This paper efficiently automates Bayesian model averaging, selection, and combination by message passing on a Forney-style factor graph with a custom mixture node. 
    Parameter and state inference, and model comparison can then be executed simultaneously using message passing with scale factors.
    This approach shortens the model design cycle and allows for the straightforward extension to hierarchical and temporal model priors to accommodate for modeling complicated time-varying processes.}
\else
    \begin{abstract}
    Bayesian state and parameter estimation have been automated effectively in a variety of probabilistic programming languages.
    The process of model comparison on the other hand, which still requires error-prone and time-consuming manual derivations, is often overlooked despite its importance.
    This paper efficiently automates Bayesian model averaging, selection, and combination by message passing on a Forney-style factor graph with a custom mixture node. 
    Parameter and state inference, and model comparison can then be executed simultaneously using message passing with scale factors.
    This approach shortens the model design cycle and allows for the straightforward extension to hierarchical and temporal model priors to accommodate for modeling complicated time-varying processes.
    \end{abstract}
\fi

\ifEntropy
    \keyword{Factor graphs; Message passing; Model averaging; Model combination; Model selection; Probabilistic inference; Scale factors}
\else
    \keywords{Factor graphs \and Message passing \and Model averaging \and Model combination \and Model selection \and Probabilistic inference \and Scale factors}
\fi

%% file: content/02-introduction.tex
\section{Introduction} \label{sec:introduction}

The famous aphorism of George Box states: ``all models are wrong, but some are useful'' \cite{box_robustness_1979}.
It is the task of statisticians and data analysts to find a model which is most useful for a given problem.
The build, compute, critique and repeat cycle \cite{blei_build_2014}, also known as Box's loop \cite{box_science_1976}, is an iterative approach for finding the most useful model.
Any efforts in shortening this design cycle increase the chances of developing more useful models, which in turn might yield more reliable predictions, more profitable returns or more efficient operations for the problem at hand.

In this paper we choose to adopt the Bayesian formalism and therefore we will specify all tasks in Box's loop as principled probabilistic inference tasks.
In addition to the well-known parameter and state inference tasks, the critique step in the design cycle is also phrased as an inference task, known as Bayesian model comparison, which automatically embodies Occam's razor \cite[Ch. 28.1]{mackay_information_2003}.
Opposed to just selecting a single model in the critique step, for different models we better quantify our confidence about which model is best, especially when data is limited \cite[Ch. 18.5.1]{koller_probabilistic_2009}.
The uncertainty arising from prior beliefs $p(m)$ over a set of models $m$ and limited observations can be naturally included through the use of Bayes' theorem
\begin{equation} \label{eq:bayes}
    p(m \midi D) = \frac{p(D\midi m) \, p(m)}{p(D)},
\end{equation}
which describes the posterior probabilities $p(m\midi D)$ as a function of model evidences $p(D\midi m)$ and where $p(D) = \sum_m p(D\midi m)p(m)$.
Starting from Bayes' rule we can obtain different comparison methods from the literature such as Bayesian model averaging \cite{hoeting_bayesian_1999}, selection, and combination \cite{monteith_turning_2011}, which we will formally introduce in Section~\ref{sec:comparison}.
We will use Bayesian model comparison as an umbrella term for these three methods throughout this paper.

The task of state and parameter estimation has been automated in a variety of tools, e.g. \cite{cox_factor_2019, bagaev_rxinfer_2023, ge_turing_2018, bingham_pyro_2018, buchner_ultranest_2021, salvatier_probabilistic_2015, carpenter_stan_2017}.
Bayesian model comparison, however, is often regarded as a separate task, whereas it submits to the same Bayesian formalism as state and parameter estimation.
A reason for overlooking the model comparison stage in a modelling task is that the computation of model evidence $p(D\midi m)$ is in most cases not automated and therefore still requires error-prone and time-consuming manual derivations, in spite of its importance and the potential data representation improvement that can achieved by for example including a Bayesian model combination stage in the modeling process \cite{monteith_turning_2011}.

This paper aims to automate the Bayesian model comparison task and is positioned between the mixture model and `gates' approaches of \cite{kamary_testing_2018} and \cite{minka_gates_2009}, respectively, which we will describe in more detail in Section~\ref{sec:related}.
Specifically, we specify the model comparison tasks as a mixture model, similarly as in \cite{kamary_testing_2018}, on a factor graph with a custom mixture node for which we derive automatable message passing update rules, which performs both parameter and state estimation, and model comparison.
These update rules generalize model comparison to arbitrary models submitting to exact inference, as the operations of the mixture node are ignorant about the adjacent subgraphs.
Additionally, we derive three common model comparison methods from literature (Bayesian model averaging, selection, and combination) using the custom mixture node. 

In short, this paper derives automated Bayesian model comparison using message passing in a factor graph.
After positioning our paper in Section~\ref{sec:related} and after reviewing factor graphs and message passing-based probabilistic inference in Section~\ref{sec:background}, we make the following contributions:
\begin{enumerate}
    \item We show that Bayesian model comparison can be performed through message passing on a graph where the individual model performances are captured in a single factor node as described in Section~\ref{sec:mixture:wvfe}.
    \item We specify a universal mixture node and derive a set of custom message passing update rules in Section~\ref{sec:mixture:node}. Performing probabilistic inference with this node in conjunction with scale factors yields different Bayesian model comparison methods.
    \item Bayesian model averaging, selection, and combination are recovered and consequently automated in Sections~\ref{sec:comparison:averaging}-\ref{sec:comparison:combination} by imposing a specific structure or local constraints on the model selection variable $m$.
\end{enumerate}
We verify our presented approach in Section~\ref{sec:experiments:verification}. We illustrate its use for models with both continuous and discrete random variables in Section~\ref{sec:experiments:validation:mixed}, after which we continue with an example of voice activity detection in Section~\ref{sec:experiments:validation:vad} where we add temporal structure on $m$.
Section~\ref{sec:discussion} discusses our approach and Section~\ref{sec:conclusion} concludes the paper.

%% file: content/03-related.tex
\section{Related work} \label{sec:related}
This section discusses related work and aims at providing a clear positioning of this paper for our contributions that follow in the upcoming sections.

The task of model comparison is widely represented in the literature \cite{fragoso_bayesian_2018}, for example concerning hypothesis testing \cite{stephan_bayesian_2009, rigoux_bayesian_2014}.
Bayesian model averaging \cite{hoeting_bayesian_1999} can be interpreted as the simplest form of model comparison that uses the Bayesian formalism to retain the first level of uncertainty in the model selection process \cite{schmitt_meta-uncertainty_2023}.
Bayesian model averaging has proven to be an effective and principled approach that converges with infinite data to the single best model in the set of candidate models \cite{minka_bayesian_2000, keller_bayesian_2018, yao_using_2018}.
When the true underlying model is not part of this set, the data is often better represented by ad hoc methods \cite{domingos_bayesian_2000}, such as ensemble methods.
In \cite{monteith_turning_2011} the idea of Bayesian model comparison is introduced, which basically performs Bayesian model averaging between mixture models comprising the candidate models, with different weights.
Another ensemble method is proposed in \cite{yao_bayesian_2022, yao_using_2018} which uses (hierarchical) stacking \cite{wolpert_stacked_1992} to construct predictive densities whose weights are data-dependent.

Automating the model design cycle \cite{blei_build_2014} under the Bayesian formalism has been the goal of many probabilistic programming languages \cite{cox_factor_2019, ge_turing_2018, bagaev_rxinfer_2023, bingham_pyro_2018, buchner_ultranest_2021, salvatier_probabilistic_2015, carpenter_stan_2017}.
This paper focuses on message passing-based approaches, which leverage the conditional independencies in the model structure for performing probabilistic inference, e.g. \cite{loeliger_introduction_2004, loeliger_factor_2007, kschischang_factor_2001, dauwels_variational_2007}, which will be formally introduced in Section~\ref{sec:background:sum-product}.
Contrary to alternative sampling-based approaches, message passing excels in modularity, speed and efficiency, especially when models submit to closed-form (variational) message computations.
Throughout this paper we follow the spirit of \cite{senoz_variational_2021}, which showed that many probabilistic inference algorithms, such as (loopy) belief propagation \cite{pearl_reverend_1982, murphy_loopy_1999}, variational message passing \cite{winn_variational_2004, dauwels_variational_2007}, expectation-maximization \cite{dauwels_expectation_2005}, expectation propagation \cite{minka_expectation_2001} can all be phrased as a constrained Bethe free energy \cite{yedidia_bethe_2001} minimization procedure.
Specifically, in Section~\ref{sec:comparison} we aim to phrase different Bayesian model comparison methods as automatable message passing algorithms.
Not only has this the potential to shorten the design cycle, but also to develop novel model comparison schemes.

The connection between (automatable) state and parameter inference, versus model comparison has been explored recently by \cite{kamary_testing_2018, keller_bayesian_2018}, who frame the problem of model comparison as a "mixture model estimation" task that is obtained by combining the individual models into a mixture model with weights representing the model selection variable.
The exposition in \cite{kamary_testing_2018, keller_bayesian_2018} is based on relatively simple examples that do not easily generalize to more complex models for the model selection variable and for the individual cluster components.
In the current paper, we aim to generalize the mixture model estimation approach by an automatable message passing-based inference framework.
Specifically, we build on the results of the recently developed scale factors \cite[Ch. 6]{reller_state-space_2013}, \cite{nguyen_efficient_2022}, which we will introduce in Section~\ref{sec:background:scalefactors}.
These scale factors support efficient tracking of local summaries of model evidences, thus enabling model comparison in generic mixture models, see Sections~\ref{sec:mixture} and \ref{sec:comparison}.

The approach we present in the current paper is also similar to the concept of `gates', introduced in \cite{minka_gates_2009}.
Gates are factor nodes that switch between mixture components with which we can derive automatable message passing procedures.
Mathematically, a gate represents a factor node of the form $f(s, m) = \prod_{k=1}^K f_k(s_k)^{m_k}$, where the model selection variable $m$ is a one-of-$K$ coded vector defined as $m_k\in\{0,1\}$ subject to $\sum_{k=1}^K m_k = 1$.
The variables $s=\bigcup_{k=1}^K s_k$.
Despite the universality of the Gates approach, the inference procedures in \cite{minka_gates_2009} focus on variational inference \cite{winn_variational_2004, winn_variational_2005, dauwels_variational_2007} and expectation propagation \cite{minka_expectation_2001}.
The mixture selection variable $m$ is then updated based on ``evidence-like quantities''.
In the variational message passing case, these quantities resemble local Bethe free energy contributions, which only take into account the performance around the gate factor node, disregarding the performance contributions of other parts in the model.
Because of the local contributions, the message passing algorithm can be very sensitive to its initialization, which has the potential to yield suboptimal inference results.

In the current paper we extend gates to models submitting to exact inference using scale factors, which allows for generalizing and automating the mixture models of \cite{kamary_testing_2018, keller_bayesian_2018}.
With these advances we can automate well-known Bayesian model comparison methods using message passing, enabling the development of novel comparison methods.

%% file: content/04-background.tex
\section{Background material} \label{sec:background}
This section aims to provide a concise review of factor graphs and message passing algorithms as we deem these concepts essential to appreciate our core contributions which we present in Sections~\ref{sec:mixture} and \ref{sec:comparison}.
This review is intentionally not exhaustive, instead we provide references to works that help to obtain a deeper understanding about the material covered here.
In Section~\ref{sec:background:factorgraphs} we introduce factor graphs as a way to visualize factorizable (probabilistic) models.
Section~\ref{sec:background:sum-product} then describes how probabilistic inference can be efficiently performed through message passing utilizing the inherent factorization of the model.
The model evidence can be tracked locally with message passing using scale factors as described in Section~\ref{sec:background:scalefactors}.
Finally, Section~\ref{sec:background:vfe} introduces the variational free energy as a bound on the model evidence.

\subsection{Forney-style factor graphs} \label{sec:background:factorgraphs}
\begin{figure}[t]
    \centering
    \input{figures/example_ffg.tikz}
    \caption{A Forney-style factor graph representation of the factorized function in \eqref{eq:ffg:example_factorization}.}
    \label{fig:methods:example_ffg}
\end{figure}
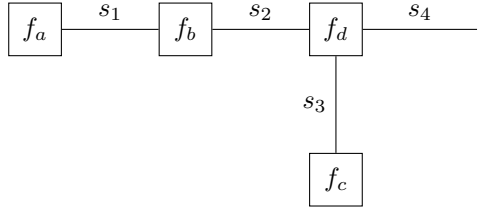

A factor graph is a specific type of probabilistic graphical model.
Here we use the Forney-style factor graph (FFG) framework as introduced in \cite{forney_codes_2001} with notational conventions adopted from \cite{loeliger_introduction_2004} to visualize our probabilistic models.
An FFG can be used to represent a factorized function
\begin{equation} \label{eq:ffg}
    f(s) =\prod_{a\in\mathcal{V}} f_a(s_a),
\end{equation}
where $s$ collects all variables in the function.
The subset $s_a \subseteq s$ contains all argument variables of a single factor $f_a$.
FFGs visualize the factorization of such a function as a graph $\mathcal{G} = (\mathcal{V}, \mathcal{E})$, where nodes $\mathcal{V}$ and edges $\mathcal{E} \subseteq \mathcal{V} \times \mathcal{V}$ represent factors and variables, respectively.
An edge connects to a node only if the variable associated with the edge is an argument of the factor associated with the node.
Nodes are indexed by the variables $a$, $b$, and $c$, where edges are indexed by $i$ and $j$ unless stated otherwise.
The set of edges connected to node $a\in\mathcal{V}$ is denoted by $\mathcal{E}(a)$ and the set of nodes connected to edge $i\in\mathcal{E}$ is referred to as $\mathcal{V}(i)$.
As an example consider the model $f(s_1,s_2,s_3,s_4)$ with factorization
\begin{equation} \label{eq:ffg:example_factorization}
    f(s_1,s_2,s_3,s_4) = f_a(s_1) f_b(s_1,s_2) f_c(s_3) f_d(s_2,s_3,s_4)
\end{equation}
The FFG representation of \eqref{eq:ffg:example_factorization} is shown in Figure~\ref{fig:methods:example_ffg}.
For a more thorough review of factor graphs, we refer the interested reader to \cite{loeliger_introduction_2004, loeliger_factor_2007}.

\subsection{Sum-product message passing} \label{sec:background:sum-product}
Consider the normalized probabilistic model
\begin{equation} \label{eq:background:model}
    p(y, s) = \prod_{a \in \mathcal{V}} f_a(y_a, s_a),
\end{equation}
with observed and latent sets of variables $y$ and $s$, respectively.
Note here that the subset $y_a\subseteq y$ can be empty, for example when dealing with prior distributions.
Upon observing the realizations $\hat{y}$, the corresponding model $p(y=\hat{y}, s)$ becomes unnormalized.
Probabilistic inference in this model then concerns the computation of the posterior distribution over the latent variables $p(s \midi y=\hat{y})$ and of the model evidence $p(y=\hat{y})$ as $p(y=\hat{y}, s) = p(s\midi y=\hat{y})p(y=\hat{y})$.
Consider the global integration over all variables in \eqref{eq:background:model} except for $s_j$ as $\int p(y=\hat{y}, s) \diff s_{\backslash j}$\footnote{Integrals are taken over the support over the variables. If a variable is discrete-valued, integral operators will be replaced with summation operators. For a consistent exposition of our work, we use integral operators throughout the paper.}.
This large global integration can be performed through a set of smaller local computations as a result of the assumed factorization in \eqref{eq:background:model}.
These smaller local computations can be considered to summarize the part of the graph that is being integrated over and are termed messages, which graphically can be interpreted to propagate over the edges in the graph.
These messages are denoted by $\mu$ and can be locally computed on the graph.
The sum-product message $\vec{\mu}_{s_j}(s_j)$ flowing out of the node $f_a(y_a =\hat{y}_a, s_a)$ with incoming messages $\vec{\mu}_{s_i}(s_i)$ is given by \cite{kschischang_factor_2001}
\begin{equation}\label{eq:sp}
    \vec{\mu}_{s_j}(s_j) = \int f_a(y_a = \hat{y}_a, s_a) \prod_{i \neq j } \vec{\mu}_{s_i}(s_i) \ \mathrm{d}s_{a\backslash j}.
\end{equation}
 We represent edges in the graph by arbitrarily directed arrows in order to distinguish between forward and backward messages propagating in or against the direction of an edge $s_j$ as $\vec{\mu}_{s_j}(s_j)$ and $\cev{\mu}_{s_j}(s_j)$, respectively.
Following this approach, the global integration reduces to the product of the messages of the variable of interest as $\int p(y=\hat{y}, s) \mathrm{d}s_{\backslash j} = \vec{\mu}_{s_j}(s_j) \cev{\mu}_{s_j}(s_j)$ for acyclic models.

Posterior distributions on edges and around nodes can then be computed according to
\begin{equation}\label{eq:marginaledge}
    p(s_j\midi y=\hat{y}) = \frac{\vec{\mu}_{s_j}(s_j)\cev{\mu}_{s_j}(s_j)}{\displaystyle \int \vec{\mu}_{s_j}(s_j)\cev{\mu}_{s_j}(s_j) \diff s_j}
\end{equation}
and
\begin{equation}\label{eq:marginalfactor}
    p(s_a\midi y=\hat{y}) = \frac{\displaystyle f_a(y_a =\hat{y}_a, s_a) \prod_{i \in \mathcal{E}(a)} \vec{\mu}_{s_i}(s_i)}{\displaystyle \int f_a(y_a =\hat{y}_a, s_a) \prod_{i \in \mathcal{E}(a)} \vec{\mu}_{s_i}(s_i) \diff s_a},
\end{equation}
respectively \cite{senoz_variational_2021}.

Derivations of the message passing update rule in \eqref{eq:sp} by phrasing inference as a constrained Bethe free energy minimization procedure are presented in \cite{senoz_variational_2021}.
Through a similar procedure one can obtain alternative message passing algorithms such as variational message passing \cite{winn_variational_2004, winn_variational_2005, dauwels_variational_2007}, expectation propagation \cite{minka_expectation_2001}, expectation maximization \cite{dauwels_expectation_2005} and hybrid algorithms.

\subsection{Scale factors}\label{sec:background:scalefactors}
The previously discussed integration $\int p(y=\hat{y},s) \diff s_{\backslash j}$ can be represented differently as
\begin{equation}
    \int p(y=\hat{y}, s) \diff s_{\backslash j} = p(y=\hat{y}) \int p(s\midi y=\hat{y}) \diff s_{\backslash j} = p(y=\hat{y}) p(s_j\midi y=\hat{y}),
\end{equation}
where $p(s_j\midi y=\hat{y})$ is the marginal distribution of $s_j$.
The implications of this result are significant: the product of two colliding sum-product messages $\vec{\mu}_{s_j}(s_j) \cev{\mu}_{s_j}(s_j)$ in an acyclic graph results in the scaled marginal distribution $p(y=\hat{y}) p(s_j\midi y=\hat{y})$.
Because of the normalization property of $p(s_j\midi y=\hat{y})$ it is possible to obtain both the normalized posterior $p(s_j\midi y=\hat{y})$ as the model evidence $p(y=\hat{y})$ on any edge and around any node in the graph. 

\begin{Theorem} \label{theorem:sf}
Consider an acyclic Forney-style factor graph $\mathcal{G}=(\mathcal{V}, \mathcal{E})$. The model evidence of the corresponding model $p(y=\hat{y},s)$ can be computed at any edge in the graph as $\int \vec{\mu}_{s_j}(s_j)\cev{\mu}_{s_j}(s_j) \diff s_j$ for all $j \in \mathcal{E}$ and at any node in the graph as $\int f_a(y_a=\hat{y}_a, s_a) \prod_{i \in \mathcal{E}(a)}\vec{\mu}_{s_i}(s_i) \diff s_a$ for all $a\in\mathcal{V}$.
\end{Theorem}
\begin{proof}
    See Appendix~\ref{appendix:proofs:sf}.
\end{proof}

What enables this local computation of the model evidence is the scaling of the messages resulting from the equality in \eqref{eq:sp}.
As a result, the messages $\vec{\mu}_{s_j}(s_j)$ can be decomposed as 
\begin{equation}
    \vec{\mu}_{s_j}(s_j) = \vec{\beta}_{s_j} \vec{p}_{s_j}(s_j),
\end{equation}
where $\vec{p}_{s_j}(s_j)$ denotes the probability distribution representing the normalized functional form of the message $\vec{\mu}_{s_j}(s_j)$. 
The term $\vec{\beta}_{s_j}$ denotes the scaling of the message $\vec{\mu}_{s_j}(s_j)$, also known as the scale factor \cite[Ch. 6]{reller_state-space_2013}, \cite{nguyen_efficient_2022}.
Scale factors can be interpreted as local summaries of the model evidence that are passed along the graph.

\subsection{Variational free energy} \label{sec:background:vfe}
In practice, however, the computation of the model evidence and therefore the posterior distribution is intractable.
Variational inference provides a generalized view that supports probabilistic inference in these types of models by approximating the exact posterior $p(s \midi y=\hat{y})$  with a variational posterior $q(s)$ that is constrained to be within a family of distributions $q\in \mathcal{Q}$.
Variational inference optimizes (the parameters of) the variational distribution $q(s)$ by minimizing the variational free energy (VFE) of a single model, defined as
\begin{equation} \label{eq:vfe}
    \mathrm{F}[q] = \mathbb{E}_{q(s)} \left[ \ln \frac{q(s)}{p(y=\hat{y},s)} \right] = \mathrm{KL}\big[q(s) \,\|\, p(s \midi y=\hat{y})\big] - \ln p(y=\hat{y}),
\end{equation}
through for example coordinate or stochastic gradient descent.

The variational free energy can serve as a bound to the model evidence in \eqref{eq:bayes} for model comparison \cite[Ch. 10.1.4]{bishop_pattern_2006}, \cite{friston_post_2011, friston_bayesian_2019}.
It is important to emphasize that the VFE not only encompasses the model evidence but also the Kullback-Leibler ($\mathrm{KL}$) divergence between the variational and exact posterior distributions obtained from the inference procedure. 

%% file: figures/example_ffg.tikz
  

    
        
        
    


\begin{tikzpicture}
  
    \node[box] (fa) {$f_a$};
    \node[box, right of=fa, node distance=20mm] (fb) {$f_b$};
    \node[box, right of=fb, node distance=20mm] (fd) {$f_d$};
    \node[box, below of=fd, node distance=20mm] (fc) {$f_c$};
    \node[right of=fd, node distance=20mm] (y) {};

    \draw[-] (fa) -- (fb)
        node[pos=0.5, above] {$s_1$};
    \draw[-] (fb) -- (fd)
        node[pos=0.5, above] {$s_2$};
    \draw[-] (fd) -- (fc)
        node[pos=0.5, left] {$s_3$};
    \draw[-] (fd) -- (y)
        node[pos=0.5, above] {$s_4$};

    
        
        
    

\end{tikzpicture}

%% file: content/05-mixture.tex
\section{Universal mixture modeling} \label{sec:mixture}
This section derives a custom factor node that allows for performing model comparison as an automatable message passing procedure in Section~\ref{sec:comparison}.
In Section~\ref{sec:mixture:wvfe} we specify a variational optimization objective for multiple models at once, where the optimization of the model selection variable can be rephrased as a probabilistic inference procedure on a separate graph with a factor node encapsulating the model-specific performance metrics.
Section~\ref{sec:mixture:node} further specifies this node and derives custom message passing update rules that allow for jointly computing \eqref{eq:bayes} and for performing state and parameter inference.

Before continuing our discussion, let us first describe the notational conventions adopted throughout this section.
In Section~\ref{sec:background} only a single model was considered.
Here we will cover $K$ normalized models, selected by the model selection variable $m$, which comprises a 1-of-$K$ binary vector with elements $m_k\in\{0, 1\}$ constrained by $\sum_{k=1}^K m_k = 1$.
The individual models $p(y_k, s_k \midi m_k=1)$ are indexed by $k$, where $y_k$ and $s_k$ collect the observed and latent variables in that model.

\subsection{A variational free energy decomposition for mixture models.} \label{sec:mixture:wvfe}

Consider the normalized joint model 
\begin{equation}\label{eq:mixture-model}
    p(y, s, m) = p(m) \prod_{k=1}^K p(y_k, s_k \midi {{m_k = 1}})^{m_k}
\end{equation}
specifying a mixture model over the individual models $p(y_k, s_k\midi{{m_k = 1}})$, with a prior $p(m)$ on the model selection variable $m$ and where $y = \bigcup_{k=1}^K y_k$ and $s=\bigcup_{k=1}^K s_k$.
Based on this joint model let us define its variational free energy $\mathrm{F}$ as
\begin{equation} \label{eq:wvfe}
\begin{split}
    \mathrm{F}[q] 
    &= \mathbb{E}_{q(s, m)}\left[\ln \frac{q(s, m)}{p(y=\hat{y}, s, m)}\right], \\
    &= \mathbb{E}_{q(m)}\left[\ln \frac{q(m)}{p(m)}\right] + \mathbb{E}_{q(m)}\left[\prod_{k=1}^K\big(\mathrm{F}_k[q]\big)^{m_k}\right],
\end{split}
\end{equation}
in which the joint variational posterior $q(s,m)$ factorizes as $q(s, m) = q(m) \prod_{k=1}^K q(s_k \mid {{m_k = 1}})^{m_k}$ and where $\mathrm{F}_k$ denotes the variational free energy of the $k^\text{th}$ model.
This decomposition is obtained by noting that $m$ is a 1-of-$K$ binary vector. Furthermore, derivations of this decomposition are provided in Appendix~\ref{appendix:derivations:wvfe}.

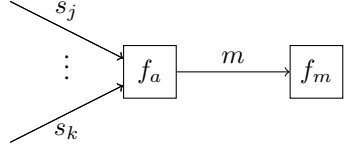
\begin{figure}
    \centering
    \input{figures/subgraph-m.tikz}
    \caption{Subgraph containing them model selection variable $m$. The node $f_m$ terminated the subgraph and is defined in \eqref{eq:factor:VFE}.}
    \label{fig:subgraph-m}
\end{figure}

This definition has also appeared in a similar form in \cite[Ch.10.1.4]{bishop_pattern_2006} and in the reinforcement learning and active inference community as an approach to policy selection \cite[Sec.2.1]{parr_generalised_2019}.
From this definition, it can be noted that the VFE for mixture models can also be written as 
\begin{equation} \label{eq:modelVFE}
    \mathrm{F}[q] = \mathbb{E}_{q(m)}\left[\ln \frac{q(m)}{p(m)f_m(m)}\right]
\end{equation}
where
\begin{equation} \label{eq:factor:VFE}
    f_m(m) = \prod_{k=1}^K \exp(-\mathrm{F}_k[q])^{m_k},
\end{equation}
as shown in Appendix~\ref{appendix:derivations:wvfe}.
This observation implies that the obtained VFEs of the individual submodels can be combined into a single factor node $f_m$, representing a scaled categorical distribution, which terminates the edge corresponding to $m$, as shown in Figure~\ref{fig:subgraph-m}.
The specification of $f_m$ allows for performing inference in the overcoupling model following existing inference procedures, similarly as in the individual submodels.
This follows in line with the validity of Bayes' theorem in \eqref{eq:bayes} for both state and parameter inference, and model comparison.
Importantly, the computation of the VFE in acyclic models can be automated \cite{senoz_variational_2021}.
Therefore, model comparison itself can also be automated.
For cyclic models, one can resort to approximating the VFE with the Bethe free energy \cite{yedidia_bethe_2001, senoz_variational_2021}.

In practice the prior model $p(m)$ might have hierarchical or temporal dynamics, including additional latent variables.
These can be incorporated without loss of generality due to the factorizable structure of the joint model, supported by the modularity of factor graphs and the corresponding message passing algorithms, as shown in Figure~\ref{fig:subgraph-m}


\subsection{A factor graph approach to universal mixture modeling: a general recipe}\label{sec:mixture:node}
In this subsection we present the general recipe for computing the posterior distributions over the variables $s$ and model selection variable $m$ in universal mixture models.
Section~\ref{sec:mixture:example} provides an illustrative example that aids the exposition in this section.
The order in which these two section are read are a matter of personal preference.

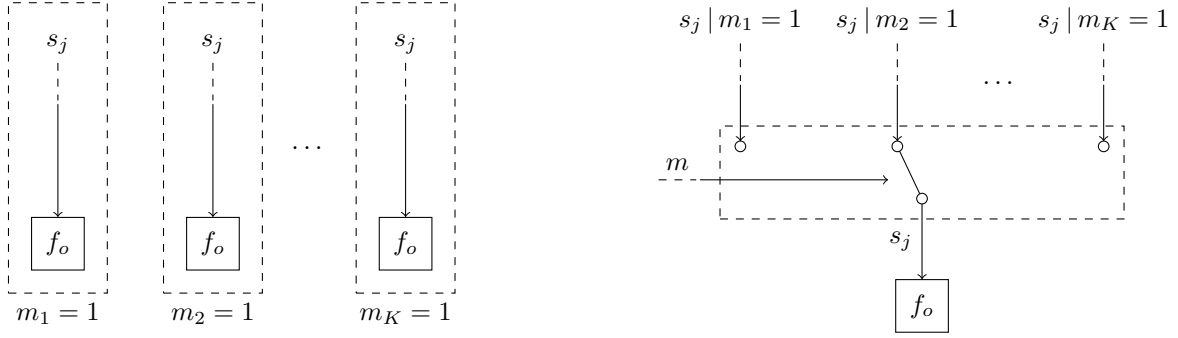
\begin{figure}[t]
     \centering
     \begin{subfigure}{0.45\linewidth}
         \centering
         \input{figures/models-separate.tikz}
         \label{fig:models:separate}
     \end{subfigure}
     \hfill
     \begin{subfigure}{0.45\linewidth}
         \centering
         \input{figures/models-joint.tikz}
         \label{fig:models:joint}
     \end{subfigure}
    \caption{(left) Overview of the traditional process of model comparison. Here inference is performed in a set of $K$ models after which the models are compared. These models may partially overlap in both variables as in structure. Specifically, in this example the variables $s_j$ connect the overlapping factors $f_o$ to the non-overlapping factors. The notation $s_j \midi {{m_k = 1}}$ denotes the variable $s_j$ in the $k^\text{th}$ model. (right) Our approach to model comparison based on mixture modeling. The different models are combined into a single graph representing a mixture model, where the model selection variable $m$ specifies the component assignment. The variable $s_j$ without conditioning implies that is has been marginalized over the different models $m$.}
    \label{fig:models}
\end{figure}

In many practical applications distinct models $p(y_k, s_k\midi {{m_k = 1}})$ partly overlap in both structure and variables.
These models may for example just differ in terms of priors or likelihood functions.
Let $f_o(y_o, s_o)$ be the product of factors which are present in all different factorizable models $p(y_k, s_k \midi {{m_k = 1}})$, with overlapping variables $s_o = \bigcap_{k=1}^K s_k$ and $y_o = \bigcap_{k=1}^K y_k$.
Based on this description we define a universal mixture model as in \cite{kamary_testing_2018} encompassing all individual models as 
\begin{align} \label{eq:fullmodel}
    p(y, s, m) &= p(m) \prod_{k=1}^K p(y_k, s_k \midi {{m_k = 1}})^{m_k} \notag \\
    &= p(m)f_o(y_o, s_o) \prod_{k=1}^K \left(\frac{p(y_k ,s_k \midi {{m_k = 1}})}{f_o(y_o, s_o)}\right)^{m_k},
\end{align}
with model selection variable $m$.
Here the overlapping factors are factored out from the mixture components.
Figure~\ref{fig:models} shows a visualization of the transformation from $K$ distinct models into a single mixture model.
With the transformation from the different models into a single mixture model presented in Figure~\ref{fig:models}, it becomes possible to include the model selection variable $m$ into the same probabilistic model.


In these universal mixture models we are often interested in computing the posterior distributions of 1) the overlapping variables $s_o$ marginalized over the distinct models and of 2) the model selection variable $m$.
Given the posterior distributions $q(s_o \midi {{m_k = 1}})$ over variables $s_o$ in a single model $m_k$, we can compute the joint posterior distribution over all overlapping variables $q(s_o)$ as
\begin{equation}\label{eq:jointposterior}
    q(s_o) = \mathbb{E}_{q(m)}\left[\prod_{k=1}^Kq(s_o \midi {{m_k = 1}})^{m_k}\right],
\end{equation}
marginalized over the different models $m$.
In the generic case, this computation follows a three-step procedure.
First, the posterior distributions $q(s_k \midi {{m_k = 1}})$ are computed in the individual submodels through an inference algorithm of choice.
Then based on the computed VFE $\mathrm{F}_k[q]$ of the individual models, the variational posterior $q(m)$ can be calculated.
Finally, the joint posterior distribution $q(s_o)$ can be computed using \eqref{eq:jointposterior}.

Here, we will restrict ourselves to acyclic submodels.
We will show that the previously described inference procedure for computing the joint posterior distributions can be performed jointly with the process of model comparison through message passing with scale factors.
In order to arrive at this point, we combine the different models into a single mixture model inspired by \cite{keller_bayesian_2018, kamary_testing_2018}.
Our specification of the mixture model, however, is more general compared to \cite{keller_bayesian_2018, kamary_testing_2018} as it does not constrain the hierarchical depth of the overlapping or distinct models and also works for nested mixture models.

Table~\ref{tab:node} introduces the novel mixture node, which acts as a switching mechanism between different models, based on the selection variable $m$.
It connects the model selection variables $m$ and the overlapping variables $s_j \midi {{m_k = 1}}$ for the different models $m_k$, to the variable $s_j$ marginalized over $m$.
Here the variables $s_j$ connect the overlapping to the non-overlapping factors.

The messages in Table~\ref{tab:node} are derived in Appendices~\ref{appendix:derivations:m} and \ref{appendix:derivations:sj} and can be intuitively understood as follows.
The message $\cev{\mu}_m(m)$ represents the unnormalized distribution over the model evidences corresponding to the individual models.
Based on the scale factors of the incoming messages, the model evidences can be computed.
The message $\cev{\mu}_{s_j\midi{{m_k = 1}}}(s_j)$ equals the incoming message from the likelihood $\cev{\mu}_{s_j}(s_j)$.
It will update $s_j\midi{{m_k = 1}}$ as if the $k^\text{th}$ model is active.
The message $\vec{\mu}_{s_j}(s_j)$ represents a mixture distribution over the incoming messages $\vec{\mu}_{s_j}(s_j)$, where the weightings are determined by the message $\vec{\mu}_m(m)$ and the scale factors of the messages $\vec{\mu}_{s_j\midi{{m_k = 1}}}(s_j)$.
This message can be propagated as a regular message over overlapping model segment yielding the marginal posterior distributions over all variables in the overlapping model segment.
\begin{Theorem}\label{theorem:node}
    Consider multiple acyclic FFGs. Given the message $\vec{\mu}_{s_j}(s_j)$ in Table~\ref{tab:node} that has been marginalized over the different models $m$. Propagating this message through the factor $f_a(y_a, s_a)$ which overlaps for all models with $s_j \in s_a$, yields again messages which are marginalized over the different models.
\end{Theorem}
\begin{proof}
    See Appendix~\ref{appendix:proofs:node}.
\end{proof}

\begin{table}[t]
    \caption{Table containing (top) the Forney-style factor graph representation of the mixture node. (bottom) The derived outgoing messages for the mixture node. It can be noted that the backward message towards $m$ resembles a scaled categorical distribution and that the forward message towards $s_j$ represents a mixture distribution. Derivations of the messages $\cev{\mu}_m(m)$ and $\vec{\mu}_{s_j}(s_j)$ are presented in Appendices~\ref{appendix:derivations:m} and \ref{appendix:derivations:sj}, respectively.} 
    \label{tab:node}
    \centering
    \input{tables/node}
\end{table}

\subsection{A factor graph approach to universal mixture modeling: an illustrative example} \label{sec:mixture:example}
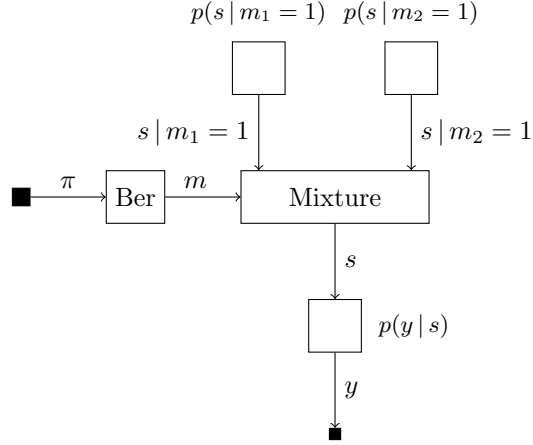
\begin{figure}
    \centering
    \input{figures/mixture-example.tikz}
    \caption{Factor graph visualization of \eqref{eq:mixture:examplejoint} in the example sketched in Section~\ref{sec:mixture:example}.}
    \label{fig:mixture:example}
\end{figure}

Consider the two probabilistic models
\begin{align}
    p(y, s \midi m_1=1) &= p(y \midi s)\, p(s \midi m_1=1), \\
    p(y, s \midi m_2=1) &= p(y \midi s)\, p(s \midi m_2=1), 
\end{align}
which share the same likelihood model with a single observed and latent variable $y$ and $s$, respectively.
The model selection variable $m$ is subject to the prior
\begin{equation}
    p(m) = \mathrm{Ber}(m \midi \pi) = \pi^{m_1}(1-\pi)^{m_2},
\end{equation}
with $\pi$ denoting the success probability.
This allows for the specification of the mixture model
\begin{equation}\label{eq:mixture:examplejoint}
    p(y, s, m) = p(m) \, p(y \midi s) \prod_{k=1}^2 p(s\midi{{m_k = 1}})^{m_k},
\end{equation}
which we visualize in Figure~\ref{fig:mixture:example}.

Suppose we are interested in computing the posterior probabilities $p(s\midi y=\hat{y})$, marginalized over the distinct models, and $p(m \midi y=\hat{y})$.
The model evidences of both models can be computed using scale factors locally on the edge corresponding to $s$ as
\begin{align*}
    p(y=\hat{y}\midi {{m_1 = 1}}) &= \int \vec{\mu}_{s\vert{{m_1 = 1}}}(s) \cev{\mu}_s(s) \diff s = \int p(s\midi{{m_1 = 1}}) \, p(y=\hat{y}\midi s) \diff s, \\
    p(y=\hat{y}\midi {{m_2 = 1}}) &= \int \vec{\mu}_{s\vert{{m_2 = 1}}}(s) \cev{\mu}_s(s) \diff s = \int p(s\midi{{m_2 = 1}}) \, p(y=\hat{y}\midi s) \diff s,
\end{align*}
which takes place inside the mixture node for computing $\cev{\mu}_m(m)$.
Together with the forward message over edge $m$, we obtain the posterior 
\begin{equation}
 p(m \midi y=\hat{y}) = \frac{\vec{\mu}_m(m)\cev{\mu}_m(m)}{\sum_{k=1}^2\vec{\mu}_m({{m_k = 1}})\cev{\mu}_m({{m_k = 1}})} = \frac{p(m) p(y=\hat{y} \midi m)}{p(y=\hat{y})}.
\end{equation}

The posterior distribution over $s$ for the first model can be computed as 
\begin{equation*}
    p(s \midi y=\hat{y}, {{m_1 = 1}}) = \frac{\vec{\mu}_{s\vert{{m_1 = 1}}}(s) \cev{\mu}_s(s)}{p(y=\hat{y} \midi {{m_1 = 1}})}.
\end{equation*}

From \eqref{eq:jointposterior} we can then compute the posterior distribution over $s$ marginalized over both models as
\begin{equation*}
    \begin{split}
        p(s \midi y=\hat{y}) 
        &= p({{m_1 = 1}}\midi y=\hat{y}) p(s \midi y=\hat{y}, {{m_1 = 1}}) \\
        &\qquad + p({{m_2 = 1}}\midi y=\hat{y}) p(s \midi y=\hat{y}, {{m_2 = 1}}), \\
        &= \frac{p({{m_1 = 1}}) p(y=\hat{y} \midi {{m_1 = 1}})}{p(y=\hat{y})} \frac{\vec{\mu}_{s\vert{{m_1 = 1}}}(s) \cev{\mu}_s(s)}{p(y=\hat{y} \midi {{m_1 = 1}})} \\
        &\qquad + \frac{p({{m_2 = 1}}) p(y=\hat{y} \midi {{m_2 = 1}})}{p(y=\hat{y})} \frac{\vec{\mu}_{s\vert{{m_2 = 1}}}(s) \cev{\mu}_s(s)}{p(y=\hat{y} \midi {{m_2 = 1}})}, \\
        &= \frac{\left(p({{m_1 = 1}})\vec{\mu}_{s\vert{{m_1 = 1}}}(s) + p({{m_2 = 1}})\vec{\mu}_{s\vert{{m_2 = 1}}}(s)\right) \cev{\mu}_s(s)}{p(y=\hat{y})}.
    \end{split}
\end{equation*}

%% file: figures/subgraph-m.tikz
\begin{tikzpicture}

    \node[box] (f) {$f_a$};
    \node[box, right = 15mm of f] (out) {$f_m$};
    \node[left = 15mm of f, yshift=10mm] (z1) {};
    \node[left = 15mm of f, yshift=-10mm] (zk) {};
    
    \draw[->] (f) -- (out)
        node[pos=0.5, above] {$m$};
    \draw[->] (z1) -- (f)
        node[pos=0.5, above] {$s_j$};
    \draw[->] (zk) -- (f)
        node[pos=0.5, below] {$s_k$};
    \draw[->] (z1) -- (f)
        node[pos=0.5] (dots1) {};
    \draw[->] (zk) -- (f)
        node[pos=0.5] (dotsk) {};
    \node (dots) at ($(dots1)!0.35!(dotsk)$) {$\vdots$};
    
\end{tikzpicture}

%% file: figures/models-separate.tikz
\begin{tikzpicture}

    \node (y1) {$s_j$};
    \node[right = 15mm of y1] (y2) {$s_j$};
    \node[right = 20mm of y2] (yK) {$s_j$};

    \node[below = 5mm of y1, inner sep=0] (y1down) {};
    \node[below = 5mm of y2, inner sep=0] (y2down) {};
    \node[below = 5mm of yK, inner sep=0] (yKdown) {};
    \node[box, below = 15mm of y1down] (y1down2) {$f_o$};
    \node[box, below = 15mm of y2down] (y2down2) {$f_o$};
    \node[box, below = 15mm of yKdown] (yKdown2) {$f_o$};

    \draw[-, dashed] (y1) -- (y1down);
    \draw[-, dashed] (y2) -- (y2down);
    \draw[-, dashed] (yK) -- (yKdown);
    \draw[->] (y1down) -- (y1down2);
    \draw[->] (y2down) -- (y2down2);
    \draw[->] (yKdown) -- (yKdown2);

    \node[fit=(y1)(y1down2), dashed, draw, inner sep=3mm, label=below:{$m_1=1$}] (boxm1) {};
    \node[fit=(y2)(y2down2), dashed, draw, inner sep=3mm, label=below:{$m_2=1$}] (boxm2) {};
    \node[fit=(yKdown2)(yK), dashed, draw, inner sep=3mm, label=below:{$m_K=1$}] (boxmK){};

    \node (dots) at ($(boxm2)!0.5!(boxmK)$) {$\dots$};

\end{tikzpicture}

%% file: figures/models-joint.tikz
\begin{tikzpicture}

    \node (y1) {$s_j \midi m_1=1$};
    \node[right = 2mm of y1] (y2) {$s_j \midi m_2=1$};
    \node[right = 8mm of y2] (yK) {$s_j \midi m_K=1$};

    \node[below = 5mm of y1, inner sep=0] (y1down) {};
    \node[below = 5mm of y2, inner sep=0] (y2down) {};
    \node[below = 5mm of yK, inner sep=0] (yKdown) {};
    \node[draw, shape = circle, below = 7.5mm of y1down, inner sep=0.5mm, outer sep = 0] (y1down2) {};
    \node[draw, shape = circle, below = 7.5mm of y2down, inner sep=0.5mm, outer sep = 0] (y2down2) {};
    \node[draw, shape = circle, below = 7.5mm of yKdown, inner sep=0.5mm, outer sep = 0] (yKdown2) {};

    \node (dots) at ($(y2down)!0.5!(yKdown)$) {$\dots$};

    \draw[-, dashed] (y1) -- (y1down);
    \draw[-, dashed] (y2) -- (y2down);
    \draw[-, dashed] (yK) -- (yKdown);
    \draw[->] (y1down) -- (y1down2);
    \draw[->] (y2down) -- (y2down2);
    \draw[->] (yKdown) -- (yKdown2);

    \node (ymiddown) at ($(y1down2)!0.5!(yKdown2)$) {};
    \node[draw, shape = circle, below = 5 mm of ymiddown, inner sep = 0.5mm, outer sep = 0] (ymiddown2) {};
    \node[box, below = 10 mm of ymiddown2, inner sep = 0] (ymiddown3) {$f_o$};

    \draw[-] (y2down2) -- (ymiddown2);
    \draw[->] (ymiddown2) -- (ymiddown3)
        node[pos=0.5, left] {$s_j$};

    \node[fit = (y1down2) (yKdown2) (ymiddown2), draw, dashed, inner sep=2mm] (xbox) {};

    \node[below = 2.5mm of y2down2] (tmp) {};
    \node[left = 25mm of tmp, inner sep = 0] (xboxright) {};
    \node[left = 5mm of xboxright, inner sep = 0] (xboxright2) {};
    
    \draw[->] (xboxright) -- (tmp);
    \draw[-, dashed] (xboxright2) -- (xboxright)
        node[pos=0.5, above] {$m$};

\end{tikzpicture}

%% file: tables/node.tex
\begin{tabular}{|c|l|}
    
    \hline 
    \multicolumn{2}{|l|}{\textbf{Factor node}} \\
    \hline
    \multicolumn{2}{|c|}{\input{figures/mixture-node.tikz}} \\
    \hline
   
    \hline
    \textbf{Messages} & \textbf{Functional form} \\
    \hline
    $\cev{\mu}_m(m)$ & $\displaystyle \prod_{k=1}^K \left(\int \vec{\mu}_{s_j\vert {{m_k=1}}}(s_j) \cev{\mu}_{s_j}(s_j) \mathrm{d}s_j\right)^{m_k}$\\
    \hline
    $\vec{\mu}_{s_j}(s_j)$ & $\displaystyle \sum_{k=1}^K \vec{\mu}_m({{m_k=1}}) \vec{\mu}_{s_j\vert {{m_k=1}}}(s_j)$ \\
    \hline
    $\cev{\mu}_{s_j\vert {{m_k=1}}}(s_j)$ & $\displaystyle \cev{\mu}_{s_j}(s_j)$ \\
    \hline
    
    \end{tabular}

%% file: figures/mixture-node.tikz
\begin{tikzpicture}

    \node (x1) {$s_j \midi {{m_1 = 1}}$};
    \node[right = 3mm of x1] (x2) {$s_j \midi {{m_2 = 1}}$};
    \node[below right = 5mm and 5mm of x2] (xdots) {$\cdots$};
    \node[above right = 5mm and 5mm of xdots] (xK) {$s_j\midi {{m_K = 1}}$};
    \node[draw, shape = circle, below = 18mm of x1, inner sep = 0.5mm, outer sep = 0] (x1-dot) {};
    \node[draw, shape = circle, below = 18mm of x2, inner sep = 0.5mm, outer sep = 0] (x2-dot) {};
    \node[draw, shape = circle, below = 18mm of xK, inner sep = 0.5mm, outer sep = 0] (xK-dot) {};
    \node[below = 2.5mm of x2-dot, inner sep = 0] (m-end) {};

    \draw[->] (x1) -- (x1-dot)
        node[pos=0.2, left] {$\downarrow$}
        node[pos=0.2, right] {\small$\vec{\mu}_{s_j\vert {{m_1 = 1}}}$}
        node[pos=0.6, left] {$\uparrow$}
        node[pos=0.6, right] {\small$\cev{\mu}_{s_j\vert {{m_1 = 1}}}$};
    \draw[->] (x2) -- (x2-dot)
        node[pos=0.2, left] {$\downarrow$}
        node[pos=0.2, right] {\small$\vec{\mu}_{s_j\vert {{m_2 = 1}}}$}
        node[pos=0.6, left] {$\uparrow$}
        node[pos=0.6, right] {\small$\cev{\mu}_{s_j\vert {{m_2 = 1}}}$};
    \draw[->] (xK) -- (xK-dot)
        node[pos=0.2, left] {$\downarrow$}
        node[pos=0.2, right] {\small$\vec{\mu}_{s_j\vert {{m_K = 1}}}$}
        node[pos=0.6, left] {$\uparrow$}
        node[pos=0.6, right] {\small$\cev{\mu}_{s_j\vert {{m_K = 1}}}$};

    \node (x-dot-mid) at ($(x1-dot)!0.5!(xK-dot)$) {};
    \node[below = 5 mm of x-dot-mid, draw, shape = circle, inner sep = 0.5mm, outer sep = 0] (y-dot) {};
    \draw[-] (x2-dot) -- (y-dot);
    \node[below = 18mm of y-dot] (y) {$s_j$}; 
    \draw[->] (y-dot) -- (y)
        node[pos=0.4, left] {$\downarrow$}
        node[pos=0.4, right] {\small$\vec{\mu}_{s_j}$}
        node[pos=0.8, left] {$\uparrow$}
        node[pos=0.8, right] {\small$\cev{\mu}_{s_j}$};

    \node[fit = (x1-dot) (xK-dot) (y-dot), draw, dashed, inner sep=3mm] (xbox) {};

    \node[left = 15mm of xbox] (m) {$m$};
    \draw[->] (m) -- (m-end)
        node[pos=0.3, above] {$\leftarrow$}
        node[pos=0.3, below] {\small$\cev{\mu}_m$}
        node[pos=0.1, above] {$\rightarrow$}
        node[pos=0.1, below] {\small$\vec{\mu}_m$};

\end{tikzpicture}

%% file: figures/mixture-example.tikz
\begin{tikzpicture}

    \node[box] (prior) {$\mathrm{Ber}$};

    \node[clamped, left=10mm of prior] (clamp) {};

    \node[box, minimum width = 25mm, right = 10mm of prior] (mixture) {$\mathrm{Mixture}$};

    \node[box, above left = 10mm and -6mm of mixture, inner sep = 0] (left) {};
    \node[above = 1mm of left] {\small$p(s\midi m_1=1)$};
    \node[box, above right = 10mm and -6mm of mixture, inner sep = 0] (right) {};
    \node[above = 1mm of right] {\small$p(s\midi m_2=1)$};

    \node[box, below = 10mm of mixture, inner sep = 0] (down) {};
    \node[clamped, below = 10mm of down, inner sep = 0] (down2) {};
    \node[right = 1mm of down] {\small$p(y\midi s)$};

    \draw[->] (clamp) -- (prior)        
        node[pos=0.5, above] {$\pi$};
    \draw[->] (prior) -- (mixture)
        node[pos=0.4, above] {$m$};
    \draw[->] (left) -- (left |- mixture.north)
        node[pos=0.5, left] {$s \midi m_1=1$};
    \draw[->] (right) -- (right |- mixture.north)
        node[pos=0.5, right] {$s \midi m_2=1$};
    \draw[->] (mixture) -- (down)
        node[pos=0.5, right] {$s$};
    \draw[->] (down) -- (down2)        
        node[pos=0.5, right] {$y$};
    
\end{tikzpicture}

%% file: content/06-comparison.tex
\section{Model comparison methods} \label{sec:comparison}
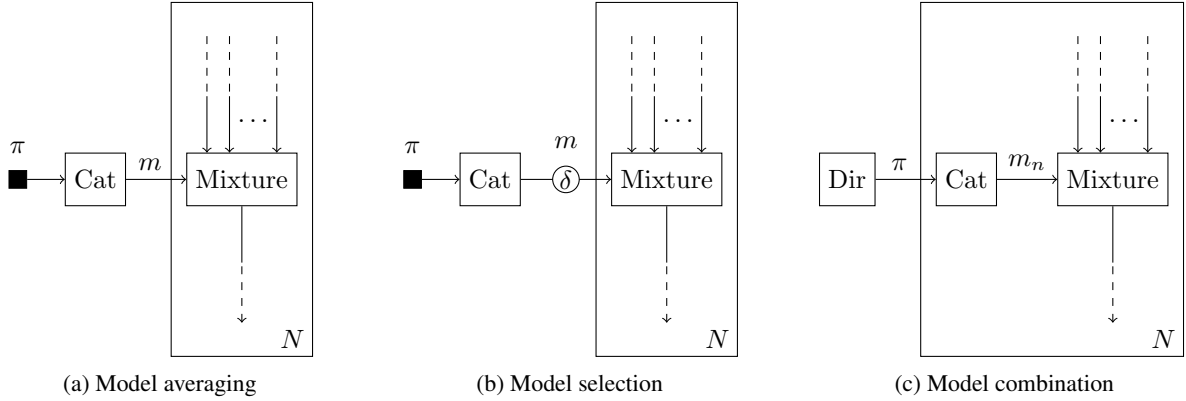
\begin{figure}[t]
    \ifEntropy
       \begin{adjustwidth}{-\extralength}{0cm}
    \fi
     \centering
     \begin{subfigure}{0.3\linewidth}
         \centering
         \input{figures/model-averaging.tikz}
         \caption{Model averaging}
         \label{fig:model-averaging}
     \end{subfigure}
     \hfill
     \begin{subfigure}{0.34\linewidth}
         \centering
         \input{figures/model-selection.tikz}
         \caption{Model selection}
         \label{fig:model-selection}
     \end{subfigure}
     \hfill
     \begin{subfigure}{0.34\linewidth}
         \centering
         \input{figures/model-combination.tikz}
         \caption{Model combination}
         \label{fig:model-combination}
     \end{subfigure}
    \caption{Schematic overview of (a) Bayesian model averaging, (b) selection and (c) combination as specified in Sections~\ref{sec:comparison:averaging}-\ref{sec:comparison:combination}. This overview explicitly visualizes the structural differences between the prior distributions and form constraints imposed on the model selection variable $m$. The edges crossing the plates are implicitly connected through equality nodes.}
    \label{fig:comparison}
    \ifEntropy
       \end{adjustwidth}
    \fi
\end{figure}

In this section, we introduce three Bayesian model comparison methods from literature: model averaging \cite{hoeting_bayesian_1999}, selection and combination \cite{monteith_turning_2011}.
For each of these methods we describe how to automate them using message passing with the mixture node in Table~\ref{tab:node}.
The factor graph approach here aids the intuitive understanding of the different approaches as their distinctions are sometimes obscure in the literature.
As we will show, each method describes an inference procedure on a slightly different model for the model selection variable $m$, possibly with different variational constraints, as visualized in Figure~\ref{fig:comparison}.

\subsection{Bayesian model averaging} \label{sec:comparison:averaging}
Bayesian model averaging (BMA) can be considered as the simplest form of model comparison and is therefore mentioned in many works, e.g. \cite{hoeting_bayesian_1999}, \cite[Ch. 14.1]{bishop_pattern_2006}.
BMA completes the model specification by specifying a categorical prior distribution over the models $m$ as
\begin{equation} \label{eq:prior-m}
    p(m) = \mathrm{Cat}(m \midi \pi),
\end{equation}
where $\pi$ denotes the vector of event probabilities.
BMA then aims at compute the posterior distribution over the models $q(m)$.
Given a set of possible models, or hypotheses, with BMA the posterior distribution $q(m)$ converges with infinite data to a Kronecker delta function that selects the single model which is the most likely given the observed set of data \cite{minka_bayesian_2000, kamary_testing_2018}.
Figure~\ref{fig:model-averaging} provides a visual representation of Bayesian model averaging.

\subsection{Bayesian model selection} \label{sec:comparison:selection}
Bayesian model selection (BMS) is a further specification of BMA as illustrated in Figure~\ref{fig:model-selection}, which selects the model out of a group of models that is the maximum a posteriori (MAP) estimate of $m$, e.g. \cite[Ch. 5.3]{murphy_machine_2012}.
Where BMA returns a posterior probability over the models $m$, BMS only returns the most probable model.
In addition to the specification of the model prior of \eqref{eq:prior-m}, BMS can be interpreted to enforce a form constraint \cite{senoz_variational_2021} on the variable $m$.
Specifically, we constrain the posterior distribution $q(m)$ to be a Kronecker delta function $\delta(\cdot)$, centered around the MAP estimate of $m$ as 
\begin{equation}
    q(m) = \delta(m - e_k),\qquad \text{s.t. }k=\argmax_k \vec{\mu}_{m}({{m_k=1}})\cev{\mu}_m({{m_k=1}}),
\end{equation}
where $e_k$ represent the $k^\text{th}$ Euclidean standard basis vector.
Figure~\ref{fig:model-selection} visualizes this constraint by the encircled $\delta$ on the edge corresponding to the variable $m$.
This form constraint will effectively interrupt the flow of the messages $\mu_m$ and instead propagate the computed marginal distribution $q(m)$ back to the connected nodes \cite[Theorem 3]{senoz_variational_2021}.
As a result $q(m)$ will be substituted for $\vec{\mu}_m(m)$ in the message $\vec{\mu}_{s_j}(s_j)$ in Table~\ref{tab:node}, which performs a selection on the incoming messages for the outgoing message as $\vec{\mu}_{s_j}(s_j) = \vec{\mu}_{s_j \vert {{m_k=1}}}(s_j)$.

\subsection{Bayesian model combination} \label{sec:comparison:combination}
Contrary to what some consider its naming implies, BMA does not find the best possible weighted set of models that explains the data and is therefore often subject to misinterpretation \cite{minka_bayesian_2000}.
Instead it performs a soft selection of the most probable model from the set of candidate models \cite{monteith_turning_2011, minka_bayesian_2000}.
With infinite data BMA converges to the single best model of the group of possible models \cite{kamary_testing_2018}.
In the case that the true model is inside the subsets of models to evaluate, this will correctly identify the true model.
However, often the true underlying model is not within this subset and therefore a suboptimal model is selected.
In this case, there might actually exist a specific weighted combination of models that represents the observed data better in terms of model evidence than the single best model \cite{minka_bayesian_2000}.

Bayesian model combination (BMC) \cite{monteith_turning_2011} has been introduced to find the best possible weighted set of models, whilst retaining uncertainty over this weighting.
The founding work of \cite{monteith_turning_2011} presents two approaches for BMC: 1) by performing an extensive search over a discretized subspace of model weightings, and 2) by sampling from a Dirichlet distribution that extends the regular categorical model prior.
Here we will illustrate the latter approach using a Dirichlet prior on $\pi$, because inference in this model can be executed efficiently using message passing.

Contrary to the previous subsection, every (set of) observation(s) is now assumed to be modeled by a distinct model $m_n$ from the set of candidate models, where $n$ indexes the observation.
Each variable $m_n$ comprises a 1-of-$K$ binary vector with elements $m_{nk}\in\{0, 1\}$ constrained by $\sum_{k=1}^K m_{nk} = 1$.
We specify the prior distribution
\begin{equation} \label{eq:prior-mn}
    p(m_n \midi \pi) = \mathrm{Cat}(m_n \midi \pi),
\end{equation}
where the event probabilities $\pi$ now appear as a random variable, which is modeled by
\begin{equation} \label{eq:prior-pi}
    p(\pi) = \mathrm{Dir}(\pi \midi \alpha),
\end{equation}
where $\alpha$ are the concentration parameters.
Intuitively, the variable $\pi$ is shared among all observations, whereas $m_n$ is specific to a single observation, as shown in Figure~\ref{fig:model-combination}.

\subsubsection{Probabilistic inference for Bayesian model combination}
Exact inference in this model is intractable because the posterior over $\pi$ resembles a mixture of Dirichlet distributions with a number of components that scales exponentially with the number of observations.
As a result, previous works in the literature have presented approximate algorithms for performing probabilistic inference in this model, such as sampling \cite{monteith_turning_2011}.
Here we will present two alternative approaches for performing approximate inference in this model.

The first approach concerns constraining the posterior distributions over $m_n$ to be Kronecker delta functions $\delta(\cdot)$ similarly as in Section~\ref{sec:comparison:selection} as 
\begin{equation}
    q(m_n) = \delta(m_n - e_k),\qquad \text{s.t. }k=\argmax_k \vec{\mu}_{m_n}(m_{nk}=1)\cev{\mu}_{m_n}(m_{nk}=1).
\end{equation}
Here we have chosen the approximate posterior $q(m_n)$ to be centered around the MAP estimate of $m_n$, however, alternative centers can also be chosen, for example by sampling from $\vec{\mu}_{m_n}(m_{nk}=1)\cev{\mu}_{m_n}(m_{nk} = 1)$.
Using this constraint the backward message $\cev{\mu}_{\pi}(\pi)$ towards $\pi$ can be computed analytically \cite[Appendix A.5]{van_de_laar_automated_2019}.
Batch or offline processing can be performed by an iterative message passing procedure, similarly as in variational message passing \cite{winn_variational_2004, winn_variational_2005, dauwels_variational_2007}, which requires initialization of the messages $\vec{\mu}_{m_n}(m_n)$ or the marginals $q(m_n)$ in order to break circular dependencies between messages and marginals in the model.
However, this approach also lends itself towards an online setting with streaming observations.
In the online setting, however, the results are heavily influenced by the prior $p(\pi)$ if chosen uninformatively as we will detail in Section~\ref{sec:experiments:verification}.
In Section~\ref{sec:experiments:verification} we also describe an approach to cope with this initialization problem.

An alternative approach for performing approximate inference in an offline manner is obtained by variational message passing \cite{winn_variational_2004, winn_variational_2005, dauwels_variational_2007}.
The true posterior distribution $p(\pi, m_1, \ldots, m_N \midi D)$ is in this case approximated by the variational posterior distribution $q(\pi, m_1, \ldots, m_N)$ being subject to a naive mean-field factorization as
\begin{equation}
    p(\pi, m_1, \ldots, m_N\midi D) \approx q(\pi, m_1, \ldots, m_N) = q(\pi)\prod_{n=1}^N q(m_n),
\end{equation}
where the individual variational distributions are constrained to have the functional forms
\begin{subequations}
    \begin{equation}
        q(\pi) = \mathrm{Dir}(\pi \midi \tilde{\alpha}),
    \end{equation}
    \begin{equation}
        q(m_n) = \mathrm{Cat}(m_n \midi \tilde{\pi}_n),
    \end{equation}
\end{subequations}
where the $\tilde{\cdot}$ accent is used to indicate the parameters of the variational posterior distributions.
Variational message passing minimizes the variational free energy by iterating the computation of variational messages and posteriors until convergence.
The corresponding variational message passing update rules are derived in \cite[Appendix A.5]{van_de_laar_automated_2019}.

%% file: figures/model-averaging.tikz
\begin{tikzpicture}

    \node[box] (prior) {$\mathrm{Cat}$};

    \node[clamped, left=5mm of prior] (clamp) {};
    \node[above=1mm of clamp] () {$\pi$};

    \node[box, right = 8mm of prior] (mixture) {$\mathrm{Mixture}$};

    \node[above left = 7.5mm and -3mm of mixture, inner sep = 0] (left) {};
    \node[above left = 7.5mm and -6mm of mixture, inner sep = 0] (left-mid) {};
    \node[above right = 4mm and -8mm of mixture, inner sep = 0] (right-mid) {$\dots$};
    \node[above right = 7.5mm and -3mm of mixture, inner sep = 0] (right) {};

    \node[below = 7.5mm of mixture, inner sep = 0] (down) {};

    \node[above = 7.5mm of left] (left2) {};
    \node[above = 7.5mm of left-mid] (left-mid2) {};
    \node[above = 7.5mm of right] (right2) {};
    \node[below = 7.5mm of down] (down2) {};

    \draw[->] (clamp) -- (prior);
    \draw[->] (prior) -- (mixture)
        node[pos=0.4, above] {$m$};
    \draw[->] (left) -- (left |- mixture.north);
    \draw[->] (left-mid) -- (left-mid |- mixture.north);
    \draw[->] (right) -- (right |- mixture.north);
    \draw[-] (mixture) -- (down);

    \draw[-, dashed] (left2) -- (left);
    \draw[-, dashed] (left-mid2) -- (left-mid);
    \draw[-, dashed] (right2) -- (right);
    \draw[->, dashed] (down) -- (down2);

    \node[draw, fit=(left2)(right2)(down2)(mixture), inner sep=2mm] (plate) {};
    \node[anchor=south east,inner sep=3pt] at (plate.south east) {$N$};
    
\end{tikzpicture}

%% file: figures/model-selection.tikz
\begin{tikzpicture}

    \node[box] (prior) {$\mathrm{Cat}$};
    
    \node[clamped, left=5mm of prior] (clamp) {};
    \node[above=1mm of clamp] () {$\pi$};

    \node[box, right = 12mm of prior] (mixture) {$\mathrm{Mixture}$};

    \node[above left = 7.5mm and -3mm of mixture, inner sep = 0] (left) {};
    \node[above left = 7.5mm and -6mm of mixture, inner sep = 0] (left-mid) {};
    \node[above right = 4mm and -8mm of mixture, inner sep = 0] (right-mid) {$\dots$};
    \node[above right = 7.5mm and -3mm of mixture, inner sep = 0] (right) {};

    \node[below = 7.5mm of mixture, inner sep = 0] (down) {};

    \node[above = 7.5mm of left] (left2) {};
    \node[above = 7.5mm of left-mid] (left-mid2) {};
    \node[above = 7.5mm of right] (right2) {};
    \node[below = 7.5mm of down] (down2) {};

    \draw[->] (clamp) -- (prior);
    \draw[->] (prior) -- (mixture)
        node[pos=0.5, circle, fill=white, draw=black, inner sep=0.2mm] {$\delta$}
        node[pos=0.5, above=3mm] {$m$};
    \draw[->] (left) -- (left |- mixture.north);
    \draw[->] (left-mid) -- (left-mid |- mixture.north);
    \draw[->] (right) -- (right |- mixture.north);
    \draw[-] (mixture) -- (down);

    \draw[-, dashed] (left2) -- (left);
    \draw[-, dashed] (left-mid2) -- (left-mid);
    \draw[-, dashed] (right2) -- (right);
    \draw[->, dashed] (down) -- (down2);

    \node[draw, fit=(left2)(right2)(down2)(mixture), inner sep=2mm] (plate) {};
    \node[anchor=south east,inner sep=3pt] at (plate.south east) {$N$};
    
\end{tikzpicture}

%% file: figures/model-combination.tikz
\begin{tikzpicture}

    \node[box] (dir) {$\mathrm{Dir}$};

    \node[box, right=8mm of dir] (cat) {$\mathrm{Cat}$};

    \node[box, right=8mm of cat] (mixture) {$\mathrm{Mixture}$};

    \node[above left = 7.5mm and -3mm of mixture, inner sep = 0] (left) {};
    \node[above left = 7.5mm and -6mm of mixture, inner sep = 0] (left-mid) {};
    \node[above right = 4mm and -8mm of mixture, inner sep = 0] (right-mid) {$\dots$};
    \node[above right = 7.5mm and -3mm of mixture, inner sep = 0] (right) {};

    \node[below = 7.5mm of mixture, inner sep = 0] (down) {};

    \node[above = 7.5mm of left] (left2) {};
    \node[above = 7.5mm of left-mid] (left-mid2) {};
    \node[above = 7.5mm of right] (right2) {};
    \node[below = 7.5mm of down] (down2) {};

    \draw[->] (dir) -- (cat)
        node[pos=0.4, above] {$\pi$};
    \draw[->] (cat) -- (mixture)
        node[pos=0.5, above] {$m_n$};
    \draw[->] (left) -- (left |- mixture.north);
    \draw[->] (left-mid) -- (left-mid |- mixture.north);
    \draw[->] (right) -- (right |- mixture.north);
    \draw[-] (mixture) -- (down);

    \draw[-, dashed] (left2) -- (left);
    \draw[-, dashed] (left-mid2) -- (left-mid);
    \draw[-, dashed] (right2) -- (right);
    \draw[->, dashed] (down) -- (down2);

    \node[draw, fit=(left2)(right2)(down2)(mixture)(cat), inner sep=2mm] (plate) {};
    \node[anchor=south east,inner sep=3pt] at (plate.south east) {$N$};
    
\end{tikzpicture}

%% file: content/07-experiments.tex
\section{Experiments} \label{sec:experiments}
In this section a set of experiments are presented for the previously presented message passing-based model comparison techniques.
Section~\ref{sec:experiments:verification} verifies the basic operations of the inference procedures for model averaging, selection and combination for data generated from a known mixture distribution.
In Section~\ref{sec:experiments:validation} the model comparison approaches are validated on application-based examples.

All experiments have been performed using the scientific programming language \texttt{Julia} \cite{bezanson_julia_2017} with the state-of-the-art probabilistic programming package \texttt{RxInfer.jl} \cite{bagaev_rxinfer_2023}.
The mixture node specified in Section~\ref{sec:mixture:node} has been integrated in its message passing engine \texttt{ReactiveMP.jl} \cite{bagaev_reactive_2023, bagaev_reactivempjl_2022}.
Aside from the results presented in the upcoming subsections, interactive \texttt{Pluto.jl} notebooks are made available online\footnote{All experiments are publicly available at \url{https://github.com/biaslab/AutomatingModelComparison}.}, allowing the reader to change hyperparameters in real-time.

\subsection{Verification experiments} \label{sec:experiments:verification}
For verification of the mixture node in Table~\ref{tab:node}, $N=\{1, 5, 10, 100, 1000\}$ observations $y_n$ have been generated from the mixture distribution
\begin{equation}\label{eq:experiments:verification}
    p(y_n) = 0.2 \,\mathcal{N}(y_n \midi -3, 1+\sigma^2) + 0.5 \,\mathcal{N}(y_n \midi 0, 1+\sigma^2) + 0.3 \,\mathcal{N}(y_n \midi 4, 1+\sigma^2),
\end{equation}
where $\mathcal{N}(y_n\midi \mu, \sigma^2)$ represents a normal distribution with mean $\mu$ and variance $\sigma^2$.
$\sigma^2$ represents the additional observation noise variance.
For the obtained data we construct the probabilistic model
\begin{subequations}
\begin{align}
    p(x_n \midi m) &= \mathcal{N}(x_n \midi -3, 1)^{m_1}\mathcal{N}(x_n \midi 0, 1)^{m_2}\mathcal{N}(x_n \midi 4, 1)^{m_3}, \\
    p(y_n\midi x_n) &= \mathcal{N}(y_n\midi x_n, \sigma^2),
\end{align}
\end{subequations}
which gets completed by the structures imposed on $m$ as introduced in Section~\ref{sec:comparison}.
Depending on the comparison method as outlined in Sections~\ref{sec:comparison:averaging}-\ref{sec:comparison:combination} we added an uninformative categorical prior on $z$ or an uninformative Dirichlet prior on the event probabilities $\pi$ that model $z$.
The aim is to infer the marginal (approximate) posterior distributions over component assignment variable $z$, for model averaging and selection, and over event probabilities $\pi$, for model combination.

For Bayesian model combination preliminary experiments showed that the results relied significantly on the initial prior $p(\pi)$ in the online setting.
Choosing this term to be uninformative, i.e. $\alpha_k << 1 \,\forall\,k$ and $\alpha_i = \alpha_j\,\forall\,i,j$, lead to a posterior distribution which became dominated by the inferred cluster assignment of the first observation.
As a result, the predictive class probability approached a delta distribution, centered around the class label of the first observation, leading to all consecutive observations being assigned to the same cluster.
This observation is as expected as the concentration parameters $\tilde{\alpha}$ of the posterior distribution $q(\pi)$ after the first model assignment $m_{1k}=1$ are updated as $\tilde{\alpha} = \alpha + m_1$, with prior concentration parameters $\alpha$.
When the entries of $\alpha$ are small, this update will have a significant effect of the updated prior distribution over $\pi$ and consecutively over the prior belief over the model assignment $\vec{\mu}_{m_n}(m_n) = \mathrm{Cat}\left(m_n \midi \alpha/\sum_{k=1}^K \alpha\right)$.
To remedy this undesirable behaviour, the prior $p(\pi)$ was chosen to prefer uniformly distributed class labels, i.e. $\alpha_k >> 1 \,\forall\,k$ and $\alpha_i = \alpha_j \,\forall\,i,j$.
Although this prior yields the same forward message $\vec{\mu}_{m_n}(m_n)$, consecutive forward messages will be less affected by the selected models $m_n$.
After the inference procedure was completed the informativeness of this prior was removed using Bayesian model reduction \cite{friston_post_2011, friston_bayesian_2019}, where the approximate posterior over $\pi$ was recomputed based on an alternative uninformative prior.

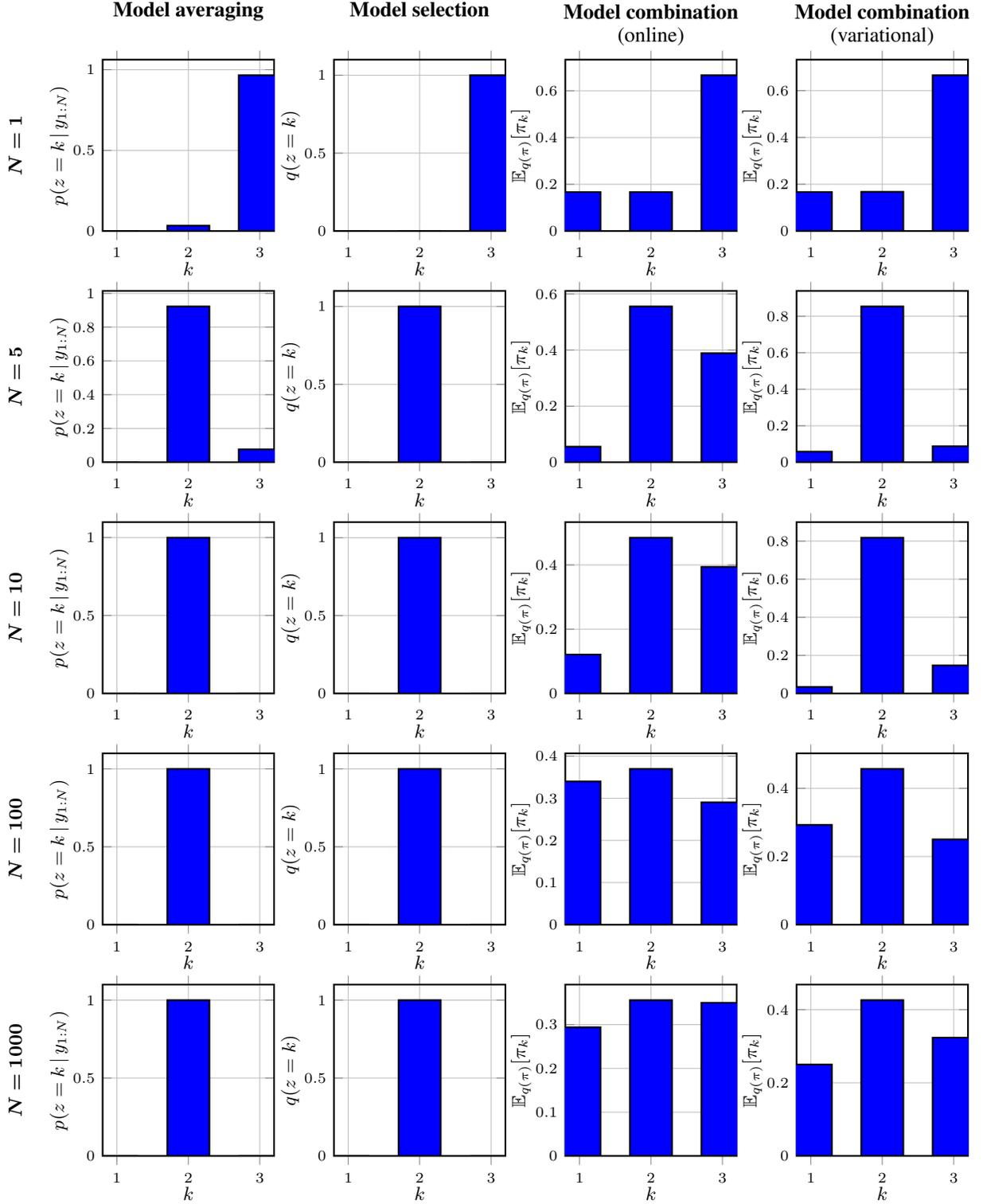
\begin{figure}
    \ifEntropy
       \begin{adjustwidth}{-\extralength}{0cm}
    \fi
    \centering
    \input{figures/experiments/verification_experiments.tikz}
    \caption{Visualization of the verification experiments as specified in Section~\ref{sec:experiments:verification}. The individual plots show the (predictive) posterior distributions for the assignment variable in \eqref{eq:experiments:verification} for $N=\{1, 5, 10, 100, 1000\}$ observations as computed using the different methods outlined in Sections~\ref{sec:comparison:averaging}-\ref{sec:comparison:combination}.}
    \label{fig:experiments:verification}
    \ifEntropy
        \end{adjustwidth}
    \fi
\end{figure}

Figure~\ref{fig:experiments:verification} shows the inferred posterior distributions of $z$ or the predictive distributions for $z$ obtained from the posterior distributions $q(\pi)$, for an observation noise variance $\sigma^2=5$.
From the results, it can be observed that Bayesian model averaging converges with increasing data to a single cluster as expected.
This selected cluster corresponds to the cluster inferred by Bayesian model selection, which also corresponds to the cluster with the highest mixing weight in \eqref{eq:experiments:verification}.
Contrary to Bayesian model selection, the alternative event probabilities obtained with Bayesian model averaging are non-zero.
Both Bayesian model combination approaches do not converge to a single cluster assignment as expected.
Instead, they better recover the data generating mixing weights in the data-generating distribution.
It can be seen that the variational approach to model combination is better capable of retrieving the original mixing weights, despite the high noise variance of $\sigma^2=5$.
The online model combination approach is less capable of retrieving the original mixing weights.
This is also as expected, single the online approach performs an approximate filtering procedure, contrary to the approximate smoothing procedure of the variational approach.
For smaller values of the noise variance, we observe in our experiments that the online model combination strategy approaches the variational strategy.

\subsection{Validation experiments} \label{sec:experiments:validation}
Aside from verifying the correctness of the message passing implementations of Section~\ref{sec:comparison} using the mixture node, this section further illustrates its usefulness in a set of validation experiments, covering real-world problems.

\subsubsection{Mixed models} \label{sec:experiments:validation:mixed}
In order to illustrate an application of the mixture node from Table~\ref{tab:node}, we show how it can be used in a mixed model where it connects continuous to discrete variables.
Consider the hypothetical situation where we wish to fit a mixture with fixed components but unknown mixing coefficients to some set of observations.
To highlight the generality of the mixture node, the mixture components are chosen to reflect shifted product distributions, where the possible shifts are limited to a discrete set of values.
The assumed probabilistic model of a single observation $y$ is given by
\begin{subequations}\label{eq:mixedmodel}
    \begin{align}
        p(a) &= \mathcal{N}(a \midi 0.5, 1),\\
        p(b) &= \mathcal{N}(b \midi 0, 1),\\
        p(c \mid z) &= \delta(c + 0.2)^{z_1} \delta(c + 1.8)^{z_2} \delta(c - 0.9)^{z_3},\\
        p(z) &= \mathrm{Cat}(z \midi 1_3/3), \\
        p(y \midi a, b, c) &= \delta(y - (ab + c)).
    \end{align}
\end{subequations}
The variables $a$, $b$ are latent variables defining the product distribution.
$c$ specifies the shift introduced on this distribution, which is picked by the selector variable $z$, comprising a 1-of-3 binary vector with elements $z_{k}\in\{0, 1\}$ constrained by $\sum_{k=1}^3 z_{k} = 1$.
$1_K$ denotes a vector of ones of length $K$.
The goal is to infer the posterior distribution of $z$ and to therefore fit this exotic mixture model to some set of data.

\begin{figure}
    \ifEntropy
       \begin{adjustwidth}{-\extralength}{0cm}
    \fi
    \centering
    \input{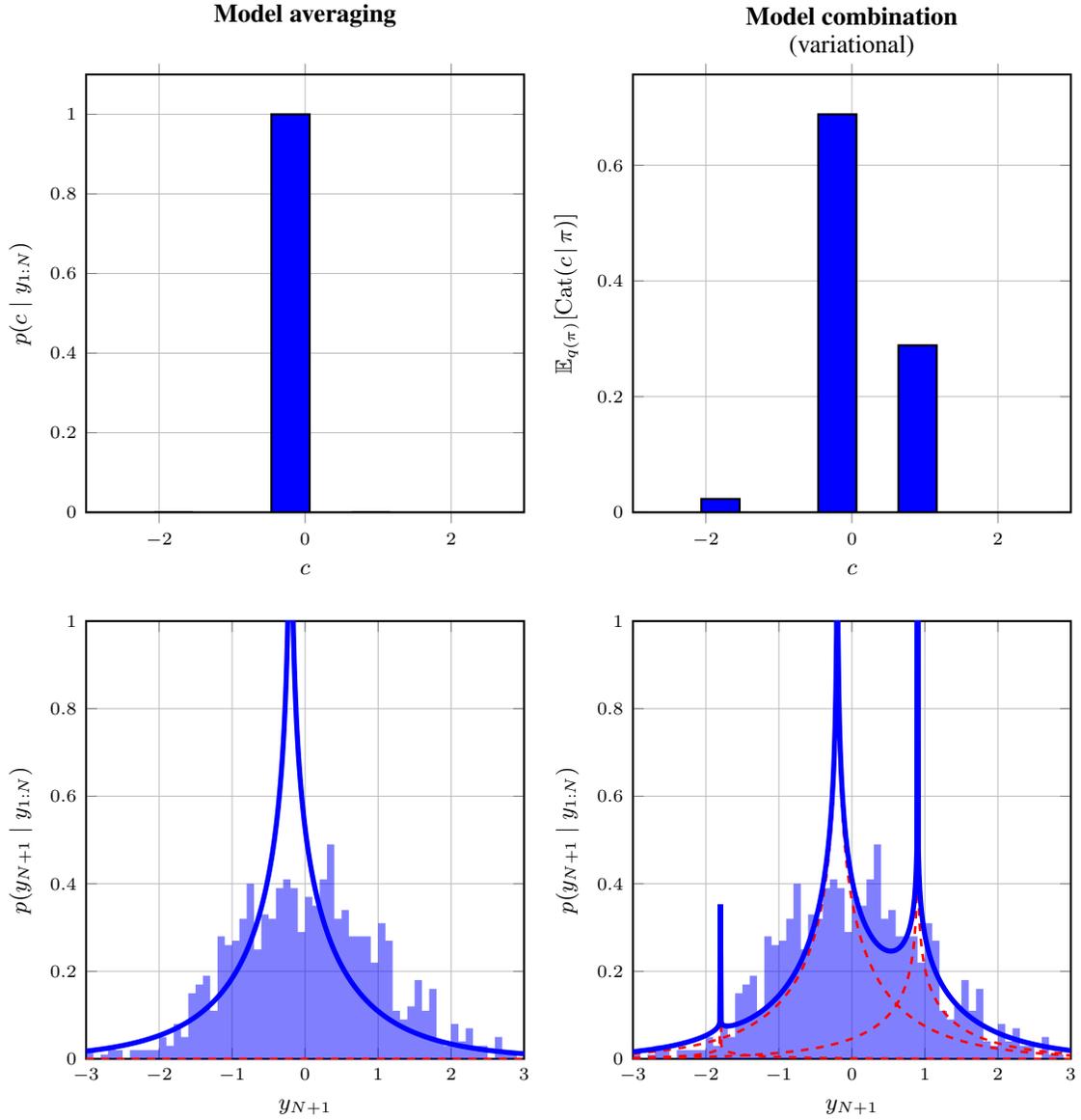}
    \caption{Inference results of the mixed model as described in Section~\ref{sec:experiments:validation:mixed}. The inference procedure is performed by (left) Bayesian model averaging and (right) Bayesian model combination under a variational mean-field factorization. (top) The posterior estimate for the shift $c$. (bottom) The predictive posterior distribution for new observations in blue with underlying components in red.}
    \label{fig:experiments:mixed}
    \ifEntropy
        \end{adjustwidth}
    \fi
\end{figure}

We perform offline probabilistic inference in this model using Bayesian model averaging and Bayesian model combination.
For the latter approach we extend the prior on $z$ with a Dirichlet distribution following Section~\ref{sec:comparison:combination} and by assuming a variational mean-field factorization.
The shifted product distributions do not yield tractable closed-form messages, therefore these distributions are approximated following \cite{cui_exact_2016}.
Figure~\ref{fig:experiments:mixed} shows the obtained data fit on a data set of 1500 observations drawn from a standard normal distribution.
This distribution does not reflect the used model in \eqref{eq:mixedmodel} on purpose to illustrate its behaviour when the true underlying model is not one of the components.
As expected model averaging converges to the most dominant component, whereas model combination attempts to improve the fit by combining the different components with fixed shifts.

\subsubsection{Voice activity detection} \label{sec:experiments:validation:vad}
In this section we illustrate a message passing approach to voice activity detection in speech that is corrupted by additive white Gaussian noise using the mixture node from Table~\ref{tab:node}.
We model speech signal $s_t$ as a first-order auto-regressive process as 
\begin{equation}
    p(s_t \midi s_{t-1}) = \mathcal{N}(s_t \midi \rho s_{t-1}, \sigma^2), 
\end{equation}
with auto-regressive parameter $\rho$ and process noise variance $\sigma^2$.
The absence of speech is modeled by independent and identically distributed variables $n_t$, which are enforced to be close to 0 as
\begin{equation}
    p(n_t) = \mathcal{N}(n_t \midi 0, 0.01).
\end{equation}
We model our observations by the mixture distribution, where we include the corruption from the additive white Gaussian noise, as
\begin{equation}
    p(y_t \midi s_t, n_t, z_t) = \mathcal{N}(y_t \midi s_t, 0.5)^{z_{t1}}    \mathcal{N}(y_t \midi n_t, 0.5)^{z_{t2}}.
\end{equation}
Here $z_t$ indicates the voice activity of the observed signal as a 1-of-2 binary vector with elements $z_{tk}\in\{0, 1\}$ constrained by $\sum_{k=1}^2 z_{tk} = 1$.
Because periods of speech are often preceded by more speech, we add temporal dynamics to $z_t$ as
\begin{equation}
    p(z_t \midi z_{t-1}) = \mathrm{Cat}(z_t \midi \mathrm{T}z_{t-1}),
\end{equation}
where the transition matrix is specified as $\mathrm{T} = [0.99999, 10^{-5}; 10^{-5}, 0.99999]$.

\begin{figure}
    \centering
    \includegraphics[width=\linewidth]{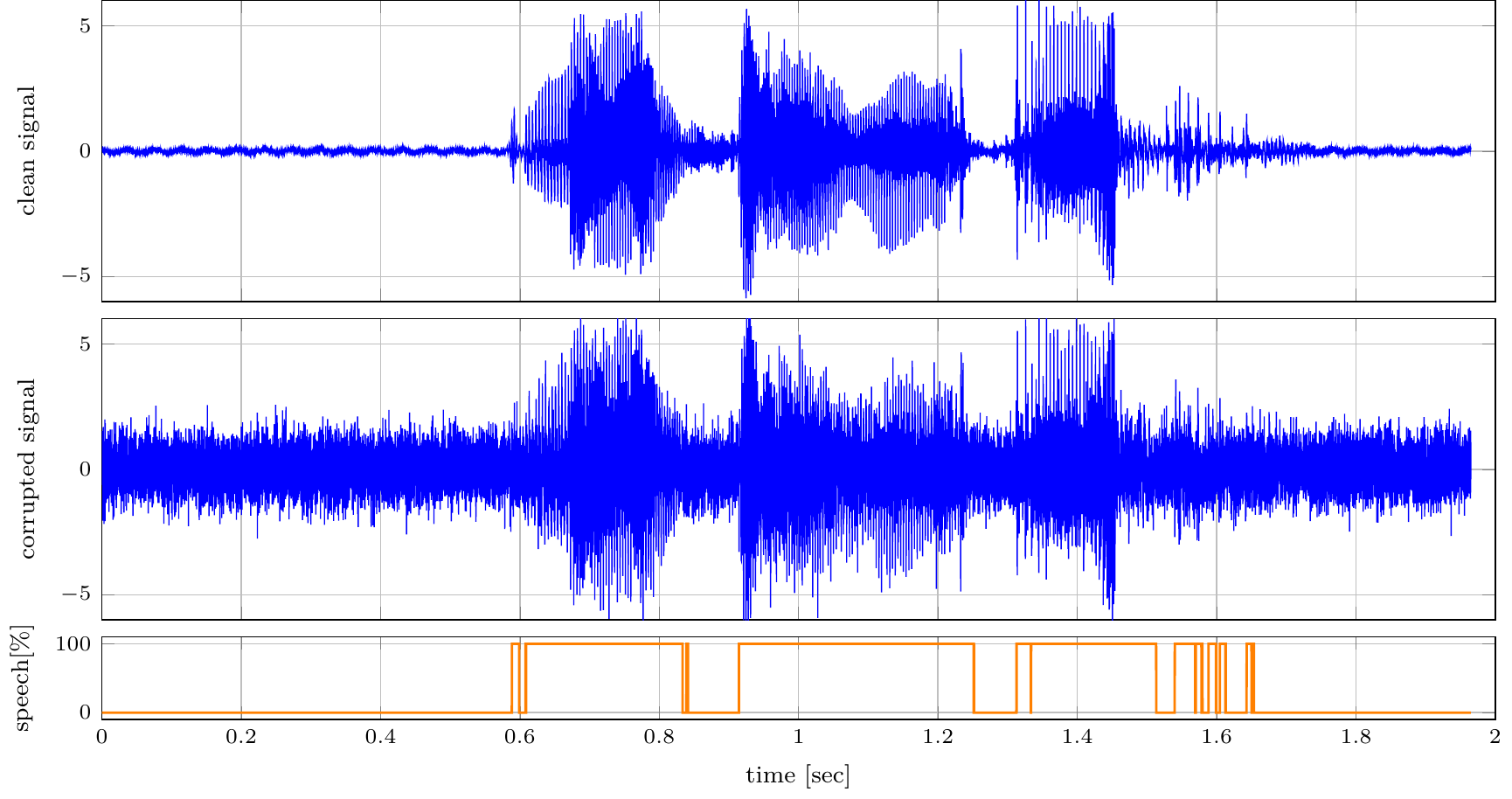}
    \caption{Results of the voice activity detection experiment as specified in Section~\ref{sec:experiments:validation:vad}. The figure shows (top) the clean signal, (middle) the clean signal corrupted by additive white Gaussian noise and (bottom) the inferred speech probability.}
    \label{fig:experiments:vad}
\end{figure}

Figure~\ref{fig:experiments:vad} shows the clean and corrupted audio signals.
The audio is sampled with a sampling frequency of 16 kHz.
The corrupted signal is used for inferring $z_t$ which is presented in the bottom plot.
Despite the corruption inflicted on the audio signal, this presented simple model is capable of detecting voice effectively as illustrated in the bottom plot of Figure~\ref{fig:experiments:vad}.

%% file: figures/experiments/verification_experiments.tikz
\begin{tikzpicture}
\begin{groupplot}[group style={group size={4 by 5}, horizontal sep={1cm}}, label style={font={\footnotesize}}, ticklabel style={font={\scriptsize}}, xtick={1,2,3}, grid={major}, ymin={0}, width={1.75in}, height={1.75in}, ylabel shift={-5pt}, xlabel shift={-5pt}, xlabel={$k$}]
    \nextgroupplot[ybar, bar width={20pt}, ylabel style={align={center}}, ylabel={$\bm{N=1}$ \\ \\ $p(z=k \,\vert\, y_{1:N})$}, style={thick}, title={\textbf{Model averaging}\\}, title style={align={center}}]
    \addplot[fill={blue}]
        table[row sep={\\}]
        {
            \\
            1.0  0.00046750498381631275  \\
            2.0  0.033487709827557874  \\
            3.0  0.9660447851886258  \\
        }
        ;
    \nextgroupplot[ybar, bar width={20pt}, ylabel={$q(z=k)$}, style={thick}, title={\textbf{Model selection}\\}, title style={align={center}}]
    \addplot[fill={blue}]
        table[row sep={\\}]
        {
            \\
            1.0  0.0  \\
            2.0  0.0  \\
            3.0  1.0  \\
        }
        ;
    \nextgroupplot[ybar, bar width={20pt}, ylabel={$\mathbb{E}_{q(\pi)}[\pi_k]$}, style={thick}, title={\textbf{Model combination}\\(online)}, title style={align={center}, yshift={-3pt}}]
    \addplot[fill={blue}]
        table[row sep={\\}]
        {
            \\
            1.0  0.1666666666666655  \\
            2.0  0.1666666666666655  \\
            3.0  0.6666666666666691  \\
        }
        ;
    \nextgroupplot[ybar, bar width={20pt}, ylabel={$\mathbb{E}_{q(\pi)}[\pi_k]$}, style={thick}, title={\textbf{Model combination}\\(variational)}, title style={align={center}, yshift={-3pt}}]
    \addplot[fill={blue}]
        table[row sep={\\}]
        {
            \\
            1.0  0.1666787185590403  \\
            2.0  0.1675451126120764  \\
            3.0  0.6657761688288834  \\
        }
        ;
    \nextgroupplot[ybar, bar width={20pt}, ylabel style={align={center}}, ylabel={$\bm{N=5}$ \\ \\ $p(z=k \,\vert\, y_{1:N})$}, style={thick}]
    \addplot[fill={blue}]
        table[row sep={\\}]
        {
            \\
            1.0  0.0009439624175522129  \\
            2.0  0.9223669629616399  \\
            3.0  0.076689074620808  \\
        }
        ;
    \nextgroupplot[ybar, bar width={20pt}, ylabel={$q(z=k)$}, style={thick}]
    \addplot[fill={blue}]
        table[row sep={\\}]
        {
            \\
            1.0  0.0  \\
            2.0  1.0  \\
            3.0  0.0  \\
        }
        ;
    \nextgroupplot[ybar, bar width={20pt}, ylabel={$\mathbb{E}_{q(\pi)}[\pi_k]$}, style={thick}]
    \addplot[fill={blue}]
        table[row sep={\\}]
        {
            \\
            1.0  0.05555555555555292  \\
            2.0  0.5555555555555577  \\
            3.0  0.3888888888888894  \\
        }
        ;
    \nextgroupplot[ybar, bar width={20pt}, ylabel={$\mathbb{E}_{q(\pi)}[\pi_k]$}, style={thick}]
    \addplot[fill={blue}]
        table[row sep={\\}]
        {
            \\
            1.0  0.05836746097587995  \\
            2.0  0.8536752773796593  \\
            3.0  0.08795726164446081  \\
        }
        ;
    \nextgroupplot[ybar, bar width={20pt}, ylabel style={align={center}}, ylabel={$\bm{N=10}$ \\ \\ $p(z=k \,\vert\, y_{1:N})$}, style={thick}]
    \addplot[fill={blue}]
        table[row sep={\\}]
        {
            \\
            1.0  3.979877322736282e-6  \\
            2.0  0.9988333595852519  \\
            3.0  0.0011626605374253835  \\
        }
        ;
    \nextgroupplot[ybar, bar width={20pt}, ylabel={$q(z=k)$}, style={thick}]
    \addplot[fill={blue}]
        table[row sep={\\}]
        {
            \\
            1.0  0.0  \\
            2.0  1.0  \\
            3.0  0.0  \\
        }
        ;
    \nextgroupplot[ybar, bar width={20pt}, ylabel={$\mathbb{E}_{q(\pi)}[\pi_k]$}, style={thick}]
    \addplot[fill={blue}]
        table[row sep={\\}]
        {
            \\
            1.0  0.12121212121212012  \\
            2.0  0.48484848484848564  \\
            3.0  0.39393939393939426  \\
        }
        ;
    \nextgroupplot[ybar, bar width={20pt}, ylabel={$\mathbb{E}_{q(\pi)}[\pi_k]$}, style={thick}]
    \addplot[fill={blue}]
        table[row sep={\\}]
        {
            \\
            1.0  0.03453399912876773  \\
            2.0  0.8179159509821997  \\
            3.0  0.14755004988903256  \\
        }
        ;
    \nextgroupplot[ybar, bar width={20pt}, ylabel style={align={center}}, ylabel={$\bm{N=100}$ \\ \\ $p(z=k \,\vert\, y_{1:N})$}, style={thick}]
    \addplot[fill={blue}]
        table[row sep={\\}]
        {
            \\
            1.0  6.385243434118563e-12  \\
            2.0  0.9999999999814935  \\
            3.0  1.2121238829254445e-11  \\
        }
        ;
    \nextgroupplot[ybar, bar width={20pt}, ylabel={$q(z=k)$}, style={thick}]
    \addplot[fill={blue}]
        table[row sep={\\}]
        {
            \\
            1.0  0.0  \\
            2.0  1.0  \\
            3.0  0.0  \\
        }
        ;
    \nextgroupplot[ybar, bar width={20pt}, ylabel={$\mathbb{E}_{q(\pi)}[\pi_k]$}, style={thick}]
    \addplot[fill={blue}]
        table[row sep={\\}]
        {
            \\
            1.0  0.33993399339934  \\
            2.0  0.36963696369636995  \\
            3.0  0.29042904290429  \\
        }
        ;
    \nextgroupplot[ybar, bar width={20pt}, ylabel={$\mathbb{E}_{q(\pi)}[\pi_k]$}, style={thick}]
    \addplot[fill={blue}]
        table[row sep={\\}]
        {
            \\
            1.0  0.2927128050247246  \\
            2.0  0.4569984684553687  \\
            3.0  0.2502887265199068  \\
        }
        ;
    \nextgroupplot[ybar, bar width={20pt}, ylabel style={align={center}}, ylabel={$\bm{N=1000}$ \\ \\ $p(z=k \,\vert\, y_{1:N})$}, style={thick}]
    \addplot[fill={blue}]
        table[row sep={\\}]
        {
            \\
            1.0  1.435619687748036e-11  \\
            2.0  0.9999999999733763  \\
            3.0  1.2267469905616448e-11  \\
        }
        ;
    \nextgroupplot[ybar, bar width={20pt}, ylabel={$q(z=k)$}, style={thick}]
    \addplot[fill={blue}]
        table[row sep={\\}]
        {
            \\
            1.0  0.0  \\
            2.0  1.0  \\
            3.0  0.0  \\
        }
        ;
    \nextgroupplot[ybar, bar width={20pt}, ylabel={$\mathbb{E}_{q(\pi)}[\pi_k]$}, style={thick}]
    \addplot[fill={blue}]
        table[row sep={\\}]
        {
            \\
            1.0  0.29403929403929346  \\
            2.0  0.3559773559773563  \\
            3.0  0.34998334998335023  \\
        }
        ;
    \nextgroupplot[ybar, bar width={20pt}, ylabel={$\mathbb{E}_{q(\pi)}[\pi_k]$}, style={thick}]
    \addplot[fill={blue}]
        table[row sep={\\}]
        {
            \\
            1.0  0.25014158986862367  \\
            2.0  0.4262591554138201  \\
            3.0  0.32359925471755624  \\
        }
        ;
\end{groupplot}
\end{tikzpicture}

%% file: content/08-discussion.tex
\section{Discussion} \label{sec:discussion}

The unifying view between probabilistic inference and model comparison as presented by this paper allows us to leverage the efficient message passing schemes for both tasks.
Interestingly, this view allows for the use of belief propagation \cite{pearl_reverend_1982}, variational message passing \cite{winn_variational_2004, winn_variational_2005, dauwels_variational_2007} and other message passing-based algorithms around the subgraph connected to the model selection variable $m$.
This insight gives rise to a novel class of model comparison algorithms, where the prior on the model selection variable is no longer constrained to be a categorical distribution, but where we now can straightforwardly introduce hierarchical and/or temporal dynamics.
Furthermore, a consequence of the automatability of the message passing algorithms is that these model comparison algorithms can easily and efficiently be implemented, without the need of error-prone and time-consuming manual derivations.

Although this paper has solely focused on message passing-based probabilistic inference, we envision interesting directions for alternative probabilistic programming packages, such as Stan \cite{carpenter_stan_2017}, Pyro \cite{bingham_pyro_2018}, Turing \cite{ge_turing_2018}, UltraNest \cite{buchner_ultranest_2021}, PyMC \cite{salvatier_probabilistic_2015}.
Currently only the PyMC framework allows for model comparison through their \texttt{compare()} function.
However, often these packages allow for estimating the (log-)evidence through sampling, or for computing the evidence lower bound (ELBO), which resembles the negative VFE of \eqref{eq:vfe}, which gets optimized using stochastic variational inference \cite{hoffman_stochastic_2013}.
An interesting direction of future research would be to use these estimates to construct the factor node $f(m)$ in \eqref{eq:factor:VFE}, with which novel model comparison algorithms can be designed, for example where the model selection variables becomes observation-dependent as in \cite{yao_bayesian_2022}.

The presented approach is especially convenient when the model allows for the use of scale factors \cite[Ch.6]{reller_state-space_2013}, \cite{nguyen_efficient_2022}.
In this way we can efficiently compute the model evidence as shown in \cite{nguyen_efficient_2022}.
The introduced mixture node in Table~\ref{tab:node} consecutively enables a simple model specification as illustrated in the source code of our experiments.

A limitation of the scale factors is that they can only be efficiently computed when the model submits to exact inference \cite{nguyen_efficient_2022}.
Extensions of the scale factors towards a variational setting would allow the use of the mixture node with a bigger variety of models.
If this limitation is resolved, then the introduced approach can be combined with more complicated models, such as for example Bayesian neural networks, whose performance is measured by the variational free energy, see e.g. \cite{blundell_weight_2015, haussmann_sampling-free_2019}.
This provides a novel solution to multi-task machine learning problems where the number of tasks is not known beforehand \cite{ruder_overview_2017}.
Each Bayesian neural network can then be trained for a specific task and additional components or networks can be added if appropriate.

The mixture nodes presented in this paper can also be nested on top of each other.
As a result, hierarchical mixture models can be realized, which can quickly increase the complexity of the nested model.
The question quickly arises where to stop.
An answer to this question is provided by Bayesian model reduction \cite{friston_post_2011, friston_bayesian_2019}.
Bayesian model reduction allows for the efficient computation of the model evidence when parts of the hierarchical model are pruned.
This approach allows for the pruning of hierarchical models in an effort to bound the complexity of the entire model.

%% file: content/09-conclusion.tex
\section{Conclusions} \label{sec:conclusion}
This paper bridges the gap between probabilistic inference for states and parameters, and for model comparison, allowing for the simultaneous automation of both tasks.
It is shown that model comparison can be performed by message passing on a graph terminated by a node that captures the performances of the different submodels, as motivated from a variational free energy perspective.
In the case where the model submits to exact inference, we can efficiently implement model comparison using our newly proposed mixture node, which leverages the efficiently computed scale factors.
Based on this node description, we show how to automate Bayesian model averaging, selection, and combination by changing the (hierarchical) prior and posterior constraints on the selection variable.

%% file: content/A1-proofs.tex
\section{Proofs} \label{appendix:proofs}

\subsection{Proof of Theorem~\ref{theorem:sf}} \label{appendix:proofs:sf}
\begin{figure}
    \centering
    \input{figures/subgraph-general.tikz}
    \caption{Visualisation of a subgraph.}
    \label{fig:subgraph-general}
\end{figure}
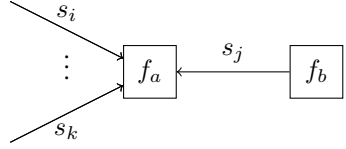

This proof follows a similar recipe as \cite[Appendix D.3]{senoz_variational_2021}.
Consider the induced subgraph in Figure~\ref{fig:subgraph-general}.
The node-local and edge-specific marginals $q_a(s_a)$ and $q_j(s_j)$, respectively, obtained by belief propagation or sum-product message passing as the fixed points of \eqref{eq:sp} are given by
\begin{subequations}
\begin{align}
    q_a(s_a) &= \frac{1}{Z_a} f_a(y=\hat{y},s_a) \prod_{i\in\mathcal{E}(a)} \vec{\mu}_{s_i}(s_i), \\
    q_j(s_j) &= \frac{1}{Z_j} \vec{\mu}_{s_j}(s_j) \cev{\mu}_{s_j}(s_j),
\end{align}
\end{subequations}
as shown in \cite[Theorem 1]{senoz_variational_2021}, with node-local and edge-specific normalization constants
\begin{subequations}
    \begin{align}
        Z_a &= \int f_a(y=\hat{y}, s_a)\prod_{i\in\mathcal{E}(a)} \vec{\mu}_{s_i}(s_i) \diff s_a, \\
        Z_j &= \int \vec{\mu}_{s_j}(s_j) \cev{\mu}_{s_j}(s_j) \diff s_j.
    \end{align}
\end{subequations}
As shown in \cite[Appendix D.3]{senoz_variational_2021} the normalization constants are equal $Z_a = Z_j$.
As this holds for all variables $s_j\in s_a$, we can deduce that $Z_i = Z_j \, \forall \, i,j\in \mathcal{E}(a)$ also holds.
Similarly, this property remains valid for adjacent nodes, allowing us to write
\begin{equation}
    Z_a = Z_j \qquad \text{s.t. } a\in\mathcal{V}, j\in\mathcal{E}.
\end{equation}
This property stipulates that the normalization constants of all (joint) marginal distributions are equal if the solutions correspond to the fixed points of \eqref{eq:sp} under sum-product message passing.
As the normalization constant equals the model evidence, we can compute the model evidence on any edge and around any node in our graph.
The fixed-point assumption is only violated in the case of cyclic graph where we perform loopy belief propagation \cite{murphy_loopy_1999}.

\subsection{Proof of Theorem~\ref{theorem:node}} \label{appendix:proofs:node}
The arbitrary outward message $\vec{\mu}_{s_k}(s_k)$ from the node $f_a(y=\hat{y}, s_a)$ can be computed using the sum-product message passing update rule in \eqref{eq:sp}.
Substitution of the definition of $\vec{\mu}_{s_j}(s_j)$ of \eqref{eq:margmessage} in this update rules yields 
\begin{equation}
    \begin{split}
        \vec{\mu}_{s_k}(s_k) 
        &= \int \vec{\mu}_{s_j}(s_j) \prod_{\substack{i\neq j \\ i\neq k}} \vec{\mu}_{s_i}(s_i) f_a(y=\hat{Y}, s_a) \diff s_{a\backslash k}, \\
        &= \int \left(\sum_{k=1}^K \vec{\mu}_m(m_k=1) \vec{\mu}_{s_j\vert m_k=1}(s_j) \right) \prod_{\substack{i\neq j \\ i\neq k}} \vec{\mu}_{s_i}(s_i) f_a(y=\hat{y}, s_a) \diff s_{a\backslash k}, \\
        & = \sum_{k=1}^K \vec{\mu}_m(m_k=1) \int \vec{\mu}_{s_j\vert m_k=1}(s_j) \prod_{\substack{i\neq j \\ i\neq k}} \vec{\mu}_{s_i}(s_i) f_a(y=\hat{y}, s_a) \diff s_{a\backslash k}, \\
        &= \sum_{k=1}^K \vec{\mu}_m(m_k=1) \vec{\mu}_{s_k\vert m_k=1}(s_k),
    \end{split}
\end{equation}
where we identify the same form as compared to $\vec{\mu}_{s_j}(s_j)$.

%% file: figures/subgraph-general.tikz
\begin{tikzpicture}

    \node[box] (f) {$f_a$};
    \node[box, right = 15mm of f] (out) {$f_b$};
    \node[left = 15mm of f, yshift=10mm] (z1) {};
    \node[left = 15mm of f, yshift=-10mm] (zk) {};
    
    \draw[->] (out) -- (f)
        node[pos=0.5, above] {$s_j$};
    \draw[->] (z1) -- (f)
        node[pos=0.5, above] {$s_i$};
    \draw[->] (zk) -- (f)
        node[pos=0.5, below] {$s_k$};
    \draw[->] (z1) -- (f)
        node[pos=0.5] (dots1) {};
    \draw[->] (zk) -- (f)
        node[pos=0.5] (dotsk) {};
    \node (dots) at ($(dots1)!0.35!(dotsk)$) {$\vdots$};
    
\end{tikzpicture}

%% file: content/A2-derivations.tex
\section{Derivations} \label{appendix:derivations}

\subsection{Derivation of variational free energy decomposition for mixture models} \label{appendix:derivations:wvfe}
Following is the derivation of \eqref{eq:wvfe}:
\begin{equation*}
\begin{split}
    \mathrm{F}[q] 
    &= \mathbb{E}_{q(s, m)}\left[\ln \frac{q(s, m)}{p(y=\hat{y}, s, m)}\right] \\
    &= \mathbb{E}_{q(s, m)}\left[\ln \frac{\displaystyle q(m) \prod_{k=1}^K q(s_k\midi {{m_k=1}})^{m_k}}{\displaystyle p(m) \prod_{k=1}^K p(y_k=\hat{y}_k, s_k \midi {{m_k=1}})^{m_k}}\right] \\    
    &= \mathbb{E}_{q(s, m)}\left[\ln \frac{q(m)}{p(m)}\right] +  \mathbb{E}_{q(s, m)}\left[ \sum_{k=1}^K m_k \ln \left(\frac{q(s_k\midi{{m_k=1}})}{p(y_k = \hat{y}_k, s_k\midi {{m_k=1}})}\right)\right] \\
    &= \mathbb{E}_{q(s, m)}\left[\ln \frac{q(m)}{p(m)}\right] +  \mathbb{E}_{q(s, m)}\left[ \prod_{k=1}^K\left( \ln \frac{q(s_k\midi{{m_k=1}})}{p(y_k=\hat{y}_k, s_k\midi {{m_k=1}})}\right)^{m_k}\right] \\
    &= \mathbb{E}_{q(m)}\left[\ln \frac{q(m)}{p(m)}\right] +  \mathbb{E}_{q(m)}\left[\prod_{k=1}^K \mathbb{E}_{q(s_k\midi{{m_k=1}})}\left[\ln \frac{q(s_k\midi{{m_k=1}})}{p(y_k=\hat{y}_k, s_k\midi {{m_k=1}})}\right]^{m_k}\right] \\
    &= \mathbb{E}_{q(m)}\left[\ln \frac{q(m)}{p(m)}\right] + \mathbb{E}_{q(m)}\left[\prod_{k=1}^K\big(\mathrm{F}_k[q]\big)^{m_k}\right]
\end{split}
\end{equation*}

The step from the third to fourth line is the result of the variable $m$ being one-hot coded.
As a result only a single $m_k$ equals $1$ and all others are constrained to be $0$.
Therefore we can obtain the identity
\begin{equation*}
    \prod_{k=1}^K a_k^{m_k} = \sum_{k=1}^K m_ka_k, \qquad \text{s.t. } \sum_{k=1}^K m_k = 1 \text{ and } m_k\in\{0, 1\} \, \forall \, k
\end{equation*}
which one might recognize as the different representations of the probability mass function of a categorical distribution.

The factor $f_m(m)$ in \eqref{eq:factor:VFE} can be derived from the above result as:
\begin{equation*}
    \begin{split}
        \mathbb{E}_{q(m)}\left[\prod_{k=1}^K\big(\mathrm{F}_k[q]\big)^{m_k}\right]
        &= \mathbb{E}_{q(m)}\left[-\ln\left(\exp\left(-\prod_{k=1}^K\big(\mathrm{F}_k[q]\big)^{m_k}\right)\right)\right] \\
        &= \mathbb{E}_{q(m)}\left[\ln \frac{1}{\exp\left(-\prod_{k=1}^K\left(\mathrm{F}_k[q]\right)^{m_k}\right)}\right] \\
        &= \mathbb{E}_{q(m)}\left[\ln \frac{1}{\prod_{k=1}^K\exp\left(-\mathrm{F}_k[q]\right)^{m_k}}\right]
    \end{split}
\end{equation*}

The step from the second to third line again uses the fact that the variable $m$ is one-hot coded.
As a result we can obtain the identity
\begin{equation*}
    \exp\left(-\prod_{k=1}^K \alpha_k^{m_k}\right) = \prod_{k=1}^K \exp(-\alpha_k)^{m_k}. \qquad \text{s.t. } \sum_{k=1}^K m_k = 1 \text{ and } m_k \in \{0, 1\} \, \forall \, k
\end{equation*}
One can validate this expression by considering a realization of $m$ and expanding the identity.

\subsection{\texorpdfstring{Derivation of message $\cev{\mu}_m(m)$}{Backward message towards m}}\label{appendix:derivations:m}
Consider the variable $s_j \in s_o$ in the context of Table~\ref{tab:node}, where we wish to compute the backwards message $\cev{\mu}_m(m)$ towards $m$. 
As shown in \cite{parr_generalised_2019} the posterior $q(m)$ can be obtained through functional optimization of \eqref{eq:wvfe}.
Following the approach stipulated in \cite{senoz_variational_2021}, the solution for $q(m)$ follows the form of \eqref{eq:marginaledge}, as 
\begin{equation}
    q(m) \propto \vec{\mu}_m(m) \cev{\mu}_m(m) = \vec{\mu}_m(m) f(m),
\end{equation}
where $\vec{\mu}_m(m)$ denotes the message towards $m$, not originating from $f(m)$.

Under the assumptions of acyclicity and tractability we obtain 
\begin{equation}\label{eq:exactF}
    \exp(-\mathrm{F}_k[q]) = Z_k = \int\vec{\mu}_{s_j\vert {{m_k=1}}}(s_j) \cev{\mu}_{s_j\vert {{m_k=1}}}(s_j) \diff s_j
\end{equation}
holds, from which we obtain the message
\begin{equation}\label{eq:message-m}
    \cev{\mu}_m(m) = f(m) = \prod_{k=1}^K\left(\int \vec{\mu}_{s_j\vert {{m_k=1}}}(s_j) \cev{\mu}_{s_j\vert {{m_k=1}}}(s_j)\diff s_j\right)^{m_k},
\end{equation}
where substitution of \eqref{eq:exactF} into the definition of $f(m)$ in \eqref{eq:factor:VFE} yields the message $\cev{\mu}_m(m)$ in Table~\ref{tab:node}.
Here we leverage scale factors inside the mixture node to compute the normalization constants of the different models.

\subsection{\texorpdfstring{Derivation of message $\vec{\mu}_{s_j}(s_j)$}{Forward message towards s}}\label{appendix:derivations:sj}
Consider again the variable $s_j \in s_o$ of which we now wish to compute its posterior distribution $q(s_j)$.
Substitution of $q(s_j\midi {{m_k=1}})$ in \eqref{eq:marginaledge} and $\cev{\mu}_m(m)$ in \eqref{eq:message-m} into \eqref{eq:jointposterior} yields
\begin{equation}
\begin{split}
    q(s_j) 
    &= \mathbb{E}_{q(m)}\left[\prod_{k=1}^K q(s_j\midi m_k)^{m_k}\right], \\
    &= \sum_{k=1}^K q(m_k) q(s_j\midi m_k), \\
    &= \sum_{k=1}^K \frac{\vec{\mu}_{m}({{m_k=1}}) \cev{\mu}_m({{m_k=1}})}{\sum_{k=1}^K \vec{\mu}_m({{m_k=1}}) \cev{\mu}_m({{m_k=1}})}  \frac{\vec{\mu}_{s_j\vert m_k}(s_j) \cev{\mu}_{s_j\vert {{m_k=1}}}(s_j)}{\int \vec{\mu}_{s_j\vert {{m_k=1}}}(s_j) \cev{\mu}_{s_j\vert {{m_k=1}}}(s_j) \diff s_j}, \\
    &= \frac{1}{\sum_{k=1}^K \vec{\mu}_m({{m_k=1}}) \cev{\mu}_m({{m_k=1}})} \sum_{k=1}^K \vec{\mu}_m({{m_k=1}}) \cev{\mu}_m({{m_k=1}}) \frac{\vec{\mu}_{s_j\vert {{m_k=1}}}(s_j) \cev{\mu}_{s_j\vert {{m_k=1}}}(s_j)}{\cev{\mu}_m({{m_k=1}})}, \\
    &= \frac{1}{Z_m} \sum_{k=1}^K \vec{\mu}_m({{m_k=1}}) \vec{\mu}_{s_j\vert {{m_k=1}}}(s_j) \cev{\mu}_{s_j\vert {{m_k=1}}}(s_j),
\end{split}
\end{equation}
where $Z_m = \sum_{k=1}^K \vec{\mu}_m({{m_k=1}})\cev{\mu}_m({{m_k=1}})$.
Furthermore, as a result of the assumption $s_j \in s_o$ being located in the overlapping model section, one of the sum-product messages towards $s_j$ is independent of the model $m_k$, i.e. either $\vec{\mu}_{s_j\vert {{m_k=1}}}(s_j) = \vec{\mu}_{s_j}(s_j)$ or $\cev{\mu}_{s_j\vert {{m_k=1}}}(s_j) = \cev{\mu}_{s_j}(s_j)$ holds.
In the situation sketched in Table~\ref{fig:models} we assume the latter.
In this case we obtain the identity
\begin{equation}\label{eq:margmessage}
    Z_m q(s_j) = \underbrace{\left(\sum_{k=1}^K \vec{\mu}_m({{m_k=1}}) \vec{\mu}_{s_j\vert {{m_k=1}}}(s_j) \right)}_{\vec{\mu}_{s_j}(s_j)} \cev{\mu}_{s_j}(s_j),
\end{equation}
from which the message $\vec{\mu}_{s_j}(s_j)$ in Table~\ref{tab:node} can be identified.